\documentclass{article}

\PassOptionsToPackage{numbers, compress}{natbib}
\usepackage[final]{nips_2017}

\usepackage[utf8]{inputenc} % allow utf-8 input
\usepackage[T1]{fontenc}    % use 8-bit T1 fonts
\usepackage{hyperref}       % hyperlinks
\usepackage{url}            % simple URL typesetting
\usepackage{booktabs}       % professional-quality tables
\usepackage{amsfonts}       % blackboard math symbols
\usepackage{nicefrac}       % compact symbols for 1/2, etc.
\usepackage{microtype}      % microtypography

\usepackage{graphicx}
\usepackage{subcaption} 
\usepackage{placeins}
\usepackage{natbib}
\usepackage{algorithm}
\usepackage{algorithmic}
\usepackage{amssymb}
\usepackage{amsmath}

\newcommand{\beq}[1][\vspace{0.3em}]{#1\begin{equation}}
\newcommand{\eeq}{\end{equation}}

\newcommand{\bit}{\vspace{0mm}\begin{itemize}}
\newcommand{\eit}{\vspace{0mm}\end{itemize}}
\newcommand{\ben}{\vspace{0mm}\begin{enumerate}}
\newcommand{\een}{\vspace{0mm}\end{enumerate}}

\newcommand{\bb}[1]{\mathbb{#1}}

\newcommand{\vect}[1]{\boldsymbol{\mathbf{#1}}}

\usepackage{color}

\title{InfoGAIL: Interpretable Imitation Learning from Visual Demonstrations}

\author{
  Yunzhu Li \\
  MIT \\
  \texttt{liyunzhu@mit.edu} \\
  \And
  Jiaming Song \\
  Stanford University \\
  \texttt{tsong@cs.stanford.edu} \\
  \And
  Stefano Ermon \\
  Stanford University \\
  \texttt{ermon@cs.stanford.edu} \\
}

\begin{document}

\maketitle

\begin{abstract}
  The goal of imitation learning is to mimic expert behavior without access to an explicit reward signal. Expert demonstrations provided by humans, however, often show significant variability due to latent factors that are typically not explicitly modeled.
  %\se{not sure this is accurate, we are modeling them}. 
  %\se{there's a logical gap between these two sentences}
  %The majority of extant works fail to address this problem and often assume that the demonstrations are from a unimodal distribution. 
  %In this paper, we propose a latent variable
  In this paper, we propose a new algorithm that can infer the latent structure of expert demonstrations in an unsupervised way. 
  Our method, built on top of Generative Adversarial Imitation Learning, can not only imitate complex behaviors, but also learn interpretable and meaningful representations of complex behavioral data, including visual demonstrations.
  %high-dimensional observations 
  %\se{representations of what?}. 
  In the driving domain, we show that a model learned from human demonstrations is able to both accurately reproduce a variety of behaviors
  %different categories of driving decisions 
 % of human-like driving behaviors 
  and accurately anticipate human actions using raw visual inputs. Compared with various baselines, our method can better capture the latent structure underlying expert demonstrations, often recovering semantically meaningful factors of variation in the data. 
  %and can imitate with respected to the recovered factors. 
  %\se{be more specific here if possible. what functionality? how big is the performance gain?}. 
  %The codes for reproducing the experiments are available at \href{https://github.com/YunzhuLi/InfoGAIL}{\texttt{https://github.com/YunzhuLi/InfoGAIL}}.
  %\se{can you fork it on the ermongroup repository? also no need for link in the abstract}
  %\yl{I will update the github readme as soon as the paper is available on arXiv}
\end{abstract}

\section{Introduction}
\label{introduction}

%The field of reinforcement learning has made significant advancements in recent years~\citep{silver2016mastering,schulman2015trust,lillicrap2015continuous,levine2013guided,schulman2015high,tamar2016value}. %\stefano{cite more folks that are likely to review this paper..}. 
%Yet, humans still significantly outperform machines in many tasks, from manipulation to driving ~\citep{pinto2016supersizing,levine2016end,chen2015deepdriving,bojarski2016end} \stefano{cite}. 

%\stefano{one difficulty is that RL require a reward function/reinforcement signal. clear in some cases, like go, but can be difficult to come up with one in complex domains, e.g., in driving}

A key limitation of reinforcement learning (RL) is that it involves the optimization of a predefined reward function or reinforcement signal~\cite{levine2013guided,schulman2015trust,lillicrap2015continuous,schulman2015high,silver2016mastering,tamar2016value}. Explicitly defining a reward function is straightforward in some cases, e.g., in games such as Go or chess.
%such as Go where the reward can be directly related with besting the opponent. 
However, designing an appropriate reward function can be difficult in more complex and less well-specified environments, e.g., for
%it could be difficult to design such a reward in complex domains where the reinforcement signal is multifaceted and specific to the environment, 
autonomous driving 
%where the vehicle has to balance between safety and efficiency.
where there is a need to balance safety, comfort, and efficiency.

Imitation learning methods have the potential to close this gap by learning how to perform tasks directly from expert demonstrations, and has succeeded in a wide range of problems ~\citep{ ziebart2008maximum,englert2015inverse,finn2016guided,Stadie2017third,ermon2015learning}.
Among them, Generative Adversarial Imitation Learning (GAIL, \cite{ho2016generative}) is a model-free imitation learning method that is highly effective and scales to relatively high dimensional environments. 
%ratliff2009learning:
The training process of GAIL can be thought of as building a generative model, which is a stochastic policy that 
%reacts to a 
when coupled with a
fixed simulation environment, produces
%data, in particular trajectories derived by actions, 
similar behaviors to the expert demonstrations. % \neal{This sentence doesn't make sense}
Similarity is achieved by jointly training
%To determine a reasonable distance metric, 
%between trajectories, 
a discriminator to distinguish expert trajectories from ones produced by the learned policy, as in GANs~\cite{goodfellow2014generative}.
%\se{cite gans}.
%this policy is jointly trained with a discriminator that distinguishes %state-action pairs from expert trajectories from fake ones. %\stefano{jointly trained with a discriminator, that distinguishes expert trajectories from fake ones}

In imitation learning, example demonstrations are typically provided by human experts. These demonstrations can show significant variability. For example, they might be collected from multiple experts, each employing a different 
%(perhaps suboptimal) 
policy. External latent factors of variation that are not explicitly captured by the simulation environment can also significantly affect the observed behavior. For example, expert demonstrations might be collected from users with different skills and habits. 
%\se{keep general. no need to mention driving at this point}.
%, and could be influenced by a large number of external factors.
%We consider learning autonomous driving through expert demonstrations, where the underlying expert policy would vary significantly from multiple human experts, since the environment, mood, driving habits, and other external factors will have impact over the experts. 
%This means that if we treat the policy from human expert trajectories as a whole, it would be an extremely complex distribution with multiple modes.
The goal of this paper is to develop an imitation learning framework that is able to automatically discover and disentangle the latent factors of variation underlying expert demonstrations.
%that are present in the expert demonstrations. 
%Under the various driving behavior patterns lies latent structures that we would like to discover and utilize. 
Analogous to the goal of uncovering style, shape, and color in generative modeling of images ~\citep{chen2016infogan}, we aim to automatically learn similar interpretable concepts from human demonstrations through an unsupervised manner. 
%\se{again, driving comes out of the blue} %This interpretable latent structure can then be used to control the high-level behavior of the policy, such as the degree of aggressiveness during driving. 
%\yunzhu{We currently aren't able to determine the degree of aggressiveness, for we only use discrete latent variables.}

We propose a new method for learning a latent variable generative model that can produce trajectories in a dynamic environment, i.e., sequences of state-actions pairs in a Markov Decision Process. Not only can the model accurately reproduce expert behavior, but also empirically learns a latent space of the observations that is semantically meaningful. Our approach is an extension of GAIL, where the objective is augmented with a mutual information term between the latent variables and the observed state-action pairs.
%For example, by learning the concept of aggressiveness from expert trajectories, we might be able to learn a policy that takes aggressiveness into account, and learn a parametrization of the space of trajectories that contains the semantic meaning of aggression.
%Moreover, users can utilize these interpretable latent structures to control the high-level behavior of the policy, such as the degree of aggressiveness during driving. 
%We further demonstrate an end-to-end imitation learning framework for discovering latent structures and learning autonomous driving from expert trajectories using raw pixel as the sole source of information.
We first illustrate the core concepts in a synthetic 2D example and then demonstrate an application in autonomous driving, where we learn to imitate complex driving behaviors while recovering semantically meaningful structure, without any supervision beyond the expert trajectories. \footnote{A video showing the experimental results is available at \href{https://youtu.be/YtNPBAW6h5k}{\texttt{https://youtu.be/YtNPBAW6h5k}}.} Remarkably, our method performs directly on raw visual inputs, using raw pixels as the only source of perceptual information. The code for reproducing the experiments are available at \href{https://github.com/ermongroup/InfoGAIL}{\texttt{https://github.com/ermongroup/InfoGAIL}}.

In particular, the contributions of this paper are threefold:
\begin{enumerate}
\itemsep0em
\item We extend GAIL with a component %that regularizes the posterior of the latent variables, through 
which
approximately maximizes the mutual information between latent space and trajectories, similar to InfoGAN \citep{chen2016infogan},
%. our model can infer the latent structure of human decision making
 resulting in a policy where low-level actions can be controlled through more abstract, high-level latent variables.
\item We extend GAIL to use raw pixels as input and produce human-like behaviors in complex high-dimensional dynamic environments.
%\item By equipping GAIL with a replay buffer and a feature extractor pre-trained on ImageNet, we allow GAIL to use raw pixel as input and produce human-like behaviors in complex high dimensional environments.
%The model can also be used to predict and categorize human behaviors very  effectively.
\item We demonstrate an application to autonomous highway driving using the TORCS driving simulator ~\citep{wymann2000torcs}. We first demonstrate that the learned policy is able to correctly navigate the track without collisions.
%navigate its way through traffic \se{is traffic the right word here?}\yl{"Morgan Freeman can now navigate your way through traffic."} without collisions. 
Then, we show that our model learns to reproduce different kinds of human-like driving behaviors by exploring the latent variable space.%, and can be used to provide accurate predictions over the continuous action space.
\end{enumerate}

%We hope that this work will inspire more real world applications to end-to-end learning approaches to self-driving using inverse reinforcement learning relying only over visual information. 

%\js{maybe talk about potential applications here, maybe later; but this is big - there are plenty of driving data in the real world, so this method has a lot of potential applications}

%\stefano{imitation learning. gail, model free, scales to relatively high dimensional models. can be thought of as building a generative model: (stochastic) policy that that combined with a fixed simulation environment, produces data that matches the expert demonstrations}
%\stefano{variability in the expert demonstrations when provided by humans. multiple experts, mood, external factors not modeled in the simulation environment. complex, multi-modal distribution}
%\stefano{would like to discover latent structure. give example. multiple drivers, more or less aggressive. we would like to learn a policy parameter that controls aggressiveness. a parameterization of the space of trajectories that is semantically meaningful.}
%\stefano{extend gail with a mutual information component, similar to infogail.}

%% What is the problem?

%% Why is it interesting and important?

%% Why is it hard? (E.g., why do naive approaches fail?)

%% Why hasn't it been solved before? (Or, what's wrong with previous proposed solutions? How does mine differ?)

%% What are the key components of my approach and results? Also include any specific limitations.

\section{Background}
\label{sec:background}
\subsection{Preliminaries}
We use the tuple $(\mathcal{S}, \mathcal{A}, P, \mathit{r}, \rho_0, \gamma)$ to define an infinite-horizon, discounted Markov decision process (MDP), where $\mathcal{S}$ represents the state space, $\mathcal{A}$ represents the action space, $P:\mathcal{S}\times\mathcal{A}\times\mathcal{S}\to\mathbb{R}$ denotes the transition probability distribution, $\mathit{r}:\mathcal{S}\to\mathbb{R}$ denotes the reward function, $\rho_0:\mathcal{S}\to\mathbb{R}$ is the distribution of the initial state $s_0$, and $\gamma\in(0,1)$ is the discount factor. Let $\pi$ denote a stochastic policy $\pi:\mathcal{S}\times\mathcal{A}\to[0,1]$, and $\pi_E$ denote the expert policy to which we only have access to demonstrations. The expert demonstrations $\tau_E$ are a set of trajectories generated using policy $\pi_E$, each of which consists of a sequence of state-action pairs. We use an expectation with respect to a policy $\pi$ to denote an expectation with respect to the trajectories it generates: $\bb{E}_\pi[f(s, a)]\triangleq\bb{E}[\sum_{t=0}^\infty\gamma^tf(s_t, a_t)]$, where $s_0 \sim \rho_0$, $a_{t} \sim \pi(a_t | s_t)$, $s_{t+1} \sim P(s_{t+1} | a_t, s_t)$.
%\footnote{We do not consider the temporal dependencies in the trajectories in this paper.}

%\stefano{we have a sequence of trajectories, sequences of state-action pairs}
%\yunzhu{Actually, we do not use the information whether two state-action pairs are from the same trajectory. Our model can work even if fragments of the trajectories are obtained. However, the dependence between state-action pairs is a trajectory-level information and can be potentially very useful.}
%\stefano{we have access to demonstrations}
%, and let $\eta(\pi)$ denote its expected discounted reward: $\eta(\pi) \triangleq\mathbb{E}_{s_0, a_0, \ldots}[\sum_{t=0}^\infty\gamma^tr(s_t,a_t)]$, where $s_0\sim\rho_0, a_t\sim\pi(a_t|s_t), s_{t+1}\sim P(s_{t+1}|s_t,a_t)$. $\pi_E$ denotes the expert policy.

\subsection{Imitation learning}
The goal of imitation learning is to learn how to perform a task directly from expert demonstrations, without any access to the reinforcement signal $r$. Typically, there are two approaches to imitation learning: 1) behavior cloning (BC), which learns a policy through supervised learning over the state-action pairs from the expert trajectories \cite{pomerleau1991efficient}; and 2) apprenticeship learning (AL), which assumes the expert policy is optimal under some unknown reward and learns a policy by recovering the reward and solving the corresponding planning problem. 
BC tends to have poor generalization properties due to compounding errors and covariate shift~\citep{ross2010efficient,ross2011reduction}. AL, on the other hand, has the advantage of learning a reward function that can be used to score trajectories~\citep{abbeel2004apprenticeship,syed2008apprenticeship,ho2016model}, but is typically expensive to run because it requires solving a reinforcement learning (RL) problem inside a learning loop.

%In the particular problem of imitation learning, the expert policy $\pi_E$ is not directly accessible, but is represented as trajectories of the states and corresponding actions pairs.
%\stefano{cost vs reward function. pick one}

%\stefano{maybe we shoud call it apprenticeship learning instead of irl}

\subsection{Generative Adversarial Imitation Learning}
%\se{keep this more general. the approach is well defined and could work even without neural networks}
Recent work on AL has adopted a different approach by learning a policy without directly estimating the corresponding reward function. 
%\stefano{the idea is to match the distribution over states-action. if that matches, then same performance. how to measure similarity between high dimensional distributions? approximate JSD with NN}
In particular, Generative Adversarial Imitation Learning (GAIL,~\cite{ho2016generative}) is a recent AL method inspired by Generative Adversarial Networks (GAN,~\cite{goodfellow2014generative}). In the GAIL framework, the agent imitates the behavior of an expert policy $\pi_E$ by matching the generated state-action distribution with the expert's distribution, where the optimum is achieved when the distance between these two distributions is minimized as measured by Jensen-Shannon divergence.
%If these two distributions match perfectly, then the agent and expert would achieve the same performance. 
%However, measuring the similarity between high-dimensional distributions is complicated, so GAIL introduces a neural network to approximately minimize the Jensen-Shannon divergence. Intuitively, the neural network is a discriminator that tries to differentiate the two distributions.   
The formal GAIL objective is denoted as %$\min_\theta \max_\omega V(\theta, \omega)$, where $V(\theta, \omega)$ is 
\beq
\label{gail:eq}
\min_{\pi}\max_{D\in (0,1)^{\mathcal{S}\times\mathcal{A}}}\bb{E}_{\pi}[\log D(s, a)] + \bb{E}_{\pi_E}[\log (1 - D(s, a))] - \lambda H(\pi)
\eeq
where $\pi$ is the policy that we wish to imitate $\pi_E$ with, $D$ is a discriminative classifier which tries to distinguish state-action pairs from the trajectories generated by $\pi$ and $\pi_E$, and $H(\pi) \triangleq \bb{E}_{\pi}[-\log \pi(a | s)]$ is the $\gamma$-discounted causal entropy of the policy $\pi_\theta$~\citep{bloem2014infinite}. %\se{is this formula correct?}\yl{I think it is correct. The expectation with respect to a policy $\pi$ is defined in Sec 2.1}. 
%We assume that our policies are Gaussian distributions with fixed standard deviations, thus $H(\pi)$ is constant in our settings.
Instead of directly learning a reward function, GAIL relies on the discriminator to guide $\pi$ into imitating the expert policy.
%\se{no need to mention specific algorithms like adam. here the story is that the objective is not differnetiable end-to-end because the simulator is a black box. so need for RL techniques (TRPO)}
%Optimization over the GAIL objective is performed by alternating between an Adam \citep{kingma2014adam} gradient step on $\omega$ to increase $V(\theta, \omega)$ with respect to $D_\omega$, and a Trust Region Policy Optimization (TRPO, \cite{schulman2015trust}) step on $\theta$ to decrease $V(\theta, \omega)$ with respect to $\pi_\theta$.

GAIL is model-free: it requires interaction with the environment to generate rollouts, but it does not need to construct a model for the environment. 
Unlike GANs, GAIL considers the environment/simulator as a black box, and thus the objective is not differentiable end-to-end.
Hence, optimization of GAIL objective requires  RL techniques based on Monte-Carlo estimation of policy gradients. Optimization over the GAIL objective is performed by alternating between a gradient step to increase (\ref{gail:eq}) with respect to the discriminator parameters, and a Trust Region Policy Optimization (TRPO,~\cite{schulman2015trust}) step to decrease (\ref{gail:eq}) with respect to $\pi$.
%\stefano{mention model-free. needs access to a simulator, but just needs rollouts}
%\stefano{I would put the infogail part first}
%\stefano{$\bb{E}_{\pi_\theta}$ undefined}
\section{Interpretable Imitation Learning through Visual Inputs}
Demonstrations are typically collected from human experts.
The resulting trajectories can show significant variability among different individuals due to internal latent factors of variation, such as levels of expertise and preferences for different strategies. Even the same individual might make different decisions while encountering the same situation, potentially resulting in demonstrations generated from multiple near-optimal but distinct policies.
%\se{think of better examples? levels of expertise? different stragies? existence of multiple near-optimal but distinct policies}
%and health conditions. 
In this section, we propose an approach that can 1) discover and disentangle salient latent factors of variation underlying expert demonstrations without supervision, 2) learn policies that produce trajectories which correspond to these latent factors, and 3) use visual inputs as the only external perceptual information.
%\stefano{it's not quite the only right? we have speed, etc. }.\js{Yunzhu says that this can be categorized as ``internal'' information; in the experiments we also treat this as auxiliary information.}

Formally, we assume that the expert policy is a mixture of experts $\pi_E = \{\pi_E^0, \pi_E^1, \dots\}$, and we define the generative process of the expert trajectory $\tau_E$ as: $s_0 \sim \rho_0$, $c \sim p(c)$, $\pi \sim p(\pi|c)$, 
$a_{t} \sim \pi(a_t | s_t)$, $s_{t+1} \sim P(s_{t+1} | a_t, s_t)$,
where $c$ is a discrete latent variable %\se{do we want to sample a fresh c at every state or not?} 
that selects a specific policy $\pi$ from the mixture of expert policies through $p(\pi|c)$ (which is unknown and needs to be learned), and $p(c)$ is the prior distribution of $c$ (which is assumed to be known before training). 
%\se{explicitly say what is known, and what is not. is $p(c)$ known? is $p(\pi|c)$ known? }
Similar to the GAIL setting, we consider the apprenticeship learning problem as the dual of an occupancy measure matching problem, and treat the trajectory $\tau_E$ as a set of state-action pairs. Instead of learning a policy solely based on the current state, we extend it to include an explicit dependence on the latent variable $c$. The objective is to recover a policy $\pi(a|s, c)$ as an approximation of $\pi_E$; when $c$ is samples from the prior $p(c)$, the trajectories $\tau$ generated by the conditional policy $\pi(a|s, c)$ should be similar to the expert trajectories $\tau_E$, as measured by a discriminative classifier. 
%\se{this problem description is too terse I think. it's an important part that needs to be explained vey well}

% where each individual expert lies within an family of probability distributions $\mc{P}_{\pi}$. The generative process of the trajectories from $\mc{P}_{\pi}$, which we denote $\{(s_t, a_t)\}_{t=0}^{\infty} \sim \mc{P}_{\pi}$, is defined as:
% $s_0 \sim \rho_0$, $\pi \sim p(\pi)$, 
% $a_{t} \sim \pi(a_t | s_t)$, $s_{t+1} \sim P(s_{t+1} | a_t, s_t)$, where $p(\pi)$ is the probability of selecting a 

% and define the generative process of the expert trajectories $\tau_E = \{(s_t, a_t)\}_{t=0}^{\infty}$ as:
% $s_0 \sim \rho_0$, $c \sim p(c)$, $\pi_c = f(c)$, 
% $a_{t} \sim \pi_c(a_t | s_t)$, $s_{t+1} \sim P(s_{t+1} | a_t, s_t)$, 
%These latent factors are independent of the specific states in the environment. 
%Hence the policy from human experts form a complex, multi-modal distribution, and the ability to capture and disentangle these factors is crucial to deal with the multi-modality. 

%\stefano{maybe try to define the problem formally. can we define a latent variable model (a POMDP) where there is a hidden state influencing the policy. define $c$ basically}

%\td{say which point corresponds to which section}

%\stefano{i think i key point to emphasize is that these latent factors are not part of the mdp state}

%\subsection{Interpretable Representation Learning}

\subsection{Interpretable Imitation Learning}
%\section{Inferring the Latent Structure of Human Decision Making}
\label{latent_structure}

Learning from demonstrations generated by a mixture of experts is challenging as we have no access to the policies employed by the individual experts. We have to 
%recover the latent structure of the expert demonstrations 
proceed 
in an unsupervised way, similar to clustering. The original Generative Adversarial Imitation Learning method would fail as it 
assumes all the demonstrations come from a single expert,
%and typically form a unimodal distribution. 
and there is no incentive in separating and disentangling variations observed in the data.
%Whereas in our case, the expert demonstrations are usually a multi-modal distribution. 
A method that can automatically disentangle the demonstrations in a meaningful way is thus needed.

The way we address this problem is to introduce a latent variable $c$ into our policy function, $\pi(a|s, c)$. Without further constraints over $c$, applying GAIL directly to this $\pi(a|s, c)$ could simply ignore $c$ and fail to separate different types of behaviors present in the expert trajectories~\footnote{For a fair comparison, we consider this form as our GAIL baseline in the experiments below.}.
%\se{Without further constraints, $\pi(a|s, c)$ could simply ignore $c$. explain what would happen if you were NOT to use info regularization. could you use gail with $c$ but without the info part?}
%To draw a connection between the latent code $c$ and the behavior of $\pi$, 
To incentivize the model to use $c$ as much as possible, we utilize an information-theoretic regularization enforcing that there should be high mutual information between $c$ and the state-action pairs in the generated trajectory. This concept was introduced by InfoGAN \citep{chen2016infogan}, where latent codes are utilized to discover the salient semantic features of the data distribution and guide the generating process. 
In particular, the regularization seeks to maximize the mutual information between latent codes and trajectories, denoted as $I(c; \tau)$,%\se{this should be between $c$ and trajectory?}, 
which is hard to maximize directly as it requires access to the posterior $P(c|\tau)$. %\se{this should be given trajectories, not pairs?}. 
Hence we introduce a variational lower bound, $L_I(\pi, Q)$, of the mutual information $I(c; \tau)$\footnote{\cite{chen2016infogan} presents a proof for the lower bound.}:
\begin{align}
L_I(\pi, Q) &= \mathbb{E}_{c\sim p(c), a\sim\pi(\cdot|s, c)}[\log Q(c|\tau)] + H(c)  \nonumber \\
%= \mathbb{E}_{a\sim\pi_\theta(\cdot|s, c)}[\mathbb{E}_{c'\sim P(c|s, a)}[\log Q(c'|s, a)]] + H(c) \\
&\le I(c;\tau)
\end{align}
where $Q(c|\tau)$ is an approximation of the true posterior $P(c|\tau)$. The objective under this regularization, which we call Information Maximizing Generative Adversarial Imitation Learning (InfoGAIL), then becomes:
\begin{equation}
\begin{split}
%\min_{\theta, \psi}\max_\omega V(\theta, \omega)  - \lambda_0\eta(\pi_\theta) - \lambda_1 L_I(\pi_\theta, Q_\psi)
\min_{\pi, Q}\max_D \bb{E}_{\pi}[\log D(s, a)] + \bb{E}_{\pi_E}[\log (1 - D(s, a))] - \lambda_1 L_I(\pi, Q) - \lambda_2 H(\pi)
\end{split}
\end{equation}
where $\lambda_1 > 0$ is the hyperparameter for information maximization regularization term, and $\lambda_2 > 0$ is the hyperparameter for the casual entropy term.
By introducing the latent code, InfoGAIL is able to identify the salient factors in the expert trajectories through mutual information maximization, and imitate the corresponding expert policy through generative adversarial training. This allows us to disentangle trajectories that may arise from a mixture of experts, such as different individuals performing the same task. 

To optimize the objective, we use a simplified posterior approximation $Q(c|s, a)$, since directly working with entire trajectories $\tau$ would be too expensive, especially when the dimension of the observations is very high (such as images). We then parameterize policy $\pi$, discriminator $D$ and posterior approximation $Q$ with weights $\theta$, $\omega$ and $\psi$ respectively. We optimize $L_I(\pi_\theta, Q_\psi)$ with stochastic gradient methods, $\pi_\theta$ using TRPO~\cite{schulman2015trust}, and $Q_\psi$ is updated using the Adam optimizer~\cite{kingma2014adam}. An outline for the optimization procedure is shown in Algorithm~\ref{alg:infogail}.
%\se{say it's in the appendix} \yl{I move the algorithm box here as per the reviewer comments.}
%The additional computational complexity is linear to the complexity of computing the inference network $Q_\psi$.

% \se{i would put optimization here}
% \se{then experiments. first synthetic. then driving. and discuss transfer in the driving? the transfer trick is pretty standard..so we can play it down a bit and save some space potentially}

\begin{algorithm}[h]
   \caption{InfoGAIL}
   \label{alg:infogail}
\begin{algorithmic}
   \STATE {\bfseries Input:} Initial parameters of policy, discriminator and posterior approximation $\theta_0, \omega_0, \psi_0$; expert trajectories $\tau_E \sim \pi_E$ containing state-action pairs.
   \STATE \textbf{Output: } Learned policy $\pi_\theta$
   \FOR{$i=0, 1, 2, ... $}
   	   \STATE Sample a batch of latent codes: $c_i \sim p(c)$
       \STATE Sample trajectories: $\tau_i\sim\pi_{\theta_i}(c_i)$, with the latent code fixed during each rollout.
       \STATE Sample state-action pairs $\chi_i \sim \tau_i$ and $\chi_E \sim \tau_E$ with same batch size. %\se{what are these? state-action pairs?}
%       \STATE Sample the same amount of state-action pairs $\chi_i$ from replay buffer, and $\chi_E$ from expert trajectories.
       \STATE Update $\omega_{i}$ to $\omega_{i+1}$ by ascending with gradients
       \vspace{-0.5em}$$\Delta_{\omega_i} = \hat{\mathbb{E}}_{\chi_i}[\nabla_{\omega_i} \log D_{\omega_i}(s, a)] + \hat{\mathbb{E}}_{\chi_E}[\nabla_{\omega_i} \log (1 - D_{\omega_i}(s, a))]$$
       \STATE Update $\psi_i$ to $\psi_{i+1}$ by descending with gradients
       \vspace{-0.5em}$$\Delta_{\psi_i} = -\lambda_1 \hat{\mathbb{E}}_{\chi_i} [\nabla_{\psi_i} \log Q_{\psi_i}(c|s, a)]$$
       \STATE Take a policy step from $\theta_i$ to $\theta_{i+1}$, using the TRPO update rule with the following objective: 
       \vspace{-0.5em}
       $$\hat{\bb{E}}_{\chi_i} [\log D_{\omega_{i+1}}(s, a)] - \lambda_1 L_I(\pi_{\theta_i}, Q_{\psi_{i+1}}) - \lambda_2 H(\pi_{\theta_i})$$ 
   \ENDFOR
\end{algorithmic}
\end{algorithm}

\subsection{Reward Augmentation}
\label{sec:rewardaugmentation}
%\td{move this topic away from driving}
%\stefano{mention it's a general way to incorporate prior knowledge.}

% \se{should we hide this part under the rug a bit and move to appendix?} \yl{I have reduced the length of the description of this point. I feel that this technique is worth mentioning as it gives us the best performance in terms of the average rollout distance.}
In complex and less well-specified environments, imitation learning methods have the potential to perform better than reinforcement learning methods as they do not require manual specification of an appropriate reward function. %beforehand in order to learn a policy. 
%Assuming the expert is following an optimal policy under some unknown reward, and that the expert is optimal, the reward function learned from the expert trajectories should match the desired reinforcement signal.
However, if the expert is performing sub-optimally, then any policy trained under the recovered rewards will be also suboptimal; in other words, the imitation learning agent's potential is bounded by the capabilities of the expert that produced the training data.
In many cases, while it is very difficult to fully specify a suitable reward function for a given task,  
%i difficult to completely specify 
it is relatively straightforward to come up with constraints that we would like to enforce over the policy. %For example, even if we cannot design an appropriate reward function capturing the subtleties of autonomous driving, we know that an autonomous vehicle should avoid colliding with other objects.

% Imitation learning methods do not require an appropriate reward function to learn a policy. 
% The existence of an expert allows us to avoid designing complex reward functions to cover every case for the environment.
% However, while imitation learning provides guidance to match the expert policy, it has no incentive to outperform the expert even if the expert is not performing under the reward function that we desire. 

This motivates the introduction of \textit{reward augmentation} \cite{englert2015inverse}, a general framework to incorporate prior knowledge in imitation learning by providing additional incentives to the agent without interfering with the imitation learning process. We achieve this by specifying a surrogate state-based reward $\eta(\pi_\theta) = \bb{E}_{s \sim \pi_\theta}[r(s)]$ that reflects our bias over the desired agent's behavior:
\begin{equation}
\min_{\theta, \psi} \max_\omega  \bb{E}_{\pi_\theta}[\log D_\omega(s, a)] + \bb{E}_{\pi_E}[\log (1 - D_\omega(s, a))] - \lambda_0 \eta(\pi_\theta) - \lambda_1 L_I(\pi_\theta, Q_\psi) - \lambda_2 H(\pi_\theta)
\end{equation}
where $\lambda_0 > 0$ is a hyperparameter. This approach can be seen as a hybrid between imitation and reinforcement learning, where part of the reinforcement signal for the policy optimization is coming from the surrogate reward and part from the discriminator, i.e., from mimicking the expert.
%The surrogate reward can also be thought of as side information provided to the generator.
%Due to the existence of an expert, the surrogate reward function can focus on certain particular aspects of the objective instead of being sophisticated enough to match every case in the environment.
For example, in our autonomous driving experiment below we show that by providing the agent with a penalty if it collides with other cars or drives off the road, we are able to significantly improve the average rollout distance of the learned policy. %This would potentially create autonomous vehicles that drivers better than humans.

\subsection{Improved Optimization}
\label{sec:optimization}

While GAIL is successful in tasks with low-dimensional inputs (in~\cite{ho2016generative}, the largest observation has 376 continuous variables), few have explored tasks where the input dimension is very high (such as images - $110 \times 200 \times 3$ pixels as in our driving experiments). In order to effectively learn a policy that relies solely on high-dimensional input, we make the following improvements over the original GAIL framework.

It is well known that the traditional GAN objective suffers from vanishing gradient and mode collapse problems~\cite{salimans2016improved,arora2017generalization}. We propose to use the Wasserstein GAN (WGAN \cite{arjovsky2017wasserstein}) technique to alleviate these problems and augment our objective function as follows:
\beq\min_{\theta, \psi}\max_{\omega}\bb{E}_{\pi_\theta}[D_\omega(s, a)] - \bb{E}_{\pi_E}[D_\omega(s, a)] - \lambda_0\eta(\pi_\theta) - \lambda_1L_I(\pi_\theta, Q_\psi) - \lambda_2 H(\pi_\theta)
\eeq
%Extending the analysis of \citep{arjovsky2017wasserstein}, it can be shown that if $D_\omega$ is $K$-Lipschitz, the objective approximately minimizes the Earth-Mover (EM) distance between the distribution of trajectories from $\pi_\theta$ and $\pi_E$. 
We note that this modification is especially important in our setting, where we want to model complex distributions over trajectories that can potentially have a large number of modes. 

We also use several variance reduction techniques, including baselines~\cite{williams1992simple} and replay buffers~\cite{mnih2015human}. Besides the baseline, we have three models to update in the InfoGAIL framework, which are represented as neural networks: the discriminator network $D_\omega(s, a)$, the policy network $\pi_\theta(a|s, c)$, and the posterior estimation network $Q_\psi(c|s, a)$. We update $D_\omega$ using RMSprop (as suggested in the original WGAN paper), and update $Q_\psi$ and $\pi_\theta$ using Adam and TRPO respectively. We include the detailed training procedure in Appendix \ref{sec:infogail-algo}.
%The training procedure is presented in Algorithm \ref{alg:infogail}. 
To speed up training, we initialize our policy from behavior cloning, as in~\cite{ho2016generative}.%\stefano{as in GAIL. i think jon did that too}

%\td{For each generator update, we update the discriminator x times.} 
Note that the discriminator network $D_\omega$ and the posterior approximation network $Q_\psi$ are treated as distinct networks, as opposed to the InfoGAN approach where they share the same network parameters until the final output layer. This is because the current WGAN training framework requires weight clipping and momentum-free optimization methods when training $D_\omega$. These changes would interfere with the training of an expressive $Q_\psi$ if $D_\omega$ and $Q_\psi$ share the same network parameters.

\section{Experiments}
\label{sec:experiments}
%\stefano{here goes the driving stuff}
%\stefano{we should keep the description of the method as general as possible. delay discussing driving as much as possible. this should go int the experiments}
We demonstrate the performance of our method by applying it first to a synthetic 2D example and then in a challenging driving domain where the agent is imitating driving behaviors from visual inputs.
By conducting experiments on these two environments, we show that our learned policy $\pi_\theta$ can 1) imitate expert behaviors using high-dimensional inputs with only a small number of expert demonstrations, 2) cluster expert behaviors into different and semantically meaningful categories, and 3) 
%behave accordingly when high level latent variables are specified.
reproduce different categories of behaviors 
by setting the high-level latent variables appropriately. 

%In most circumstances, humans are able to drive reliably without senses other than vision ~\citep{booher1978effects}, whereas present autonomous vehicles require expensive apparatus to provide much more information that humans receive. Examples include precise distance measurements from Lidars, which typically cost tens of thousands of US dollars.  

%\subsection{Environment Setup}
%The Open Racing Car Simulator (TORCS, \cite{wymann2000torcs}) is a popular simulator environment for research in autonomous vehicles. We packaged it into a client-server framework with APIs similar to OpenAI Gym \citep{brockman2016openai}. Our framework produces a realistic dashboard view and driving related information, and communicates with the policy (client) through TCP packets, so that the policy can be written in languages other than C++. In particular, we implemented our policy using the TensorFlow Python API ~\citep{abadi2016tensorflow}. %This framework and the code for reproducing the experiments are available at \hyperref[https://github.com/YunzhuLi/InfoGAIL]{\texttt{https://github.com/YunzhuLi/InfoGAIL}}.

%\textbf{Environment Setup:} 
The driving experiments are conducted in the TORCS (The Open Source Racing Car Simulator, \cite{wymann2000torcs}) environment. The demonstrations are collected by manually driving along the race track, and show typical behaviors like staying within lanes, avoiding collisions and surpassing other cars. The policy accepts raw visual inputs as the only external inputs for the state, and produces a three-dimensional continuous action that consists of \textit{steering}, \textit{acceleration}, and \textit{braking}.
We assume that our policies are Gaussian distributions with fixed standard deviations, thus $H(\pi)$ is constant.
%\footnote{Acceleration and braking can be triggered simultaneously, although there is little reason to do so.}.

\begin{figure}
\centering
\begin{subfigure}{0.23\textwidth}
\includegraphics[width=\textwidth]{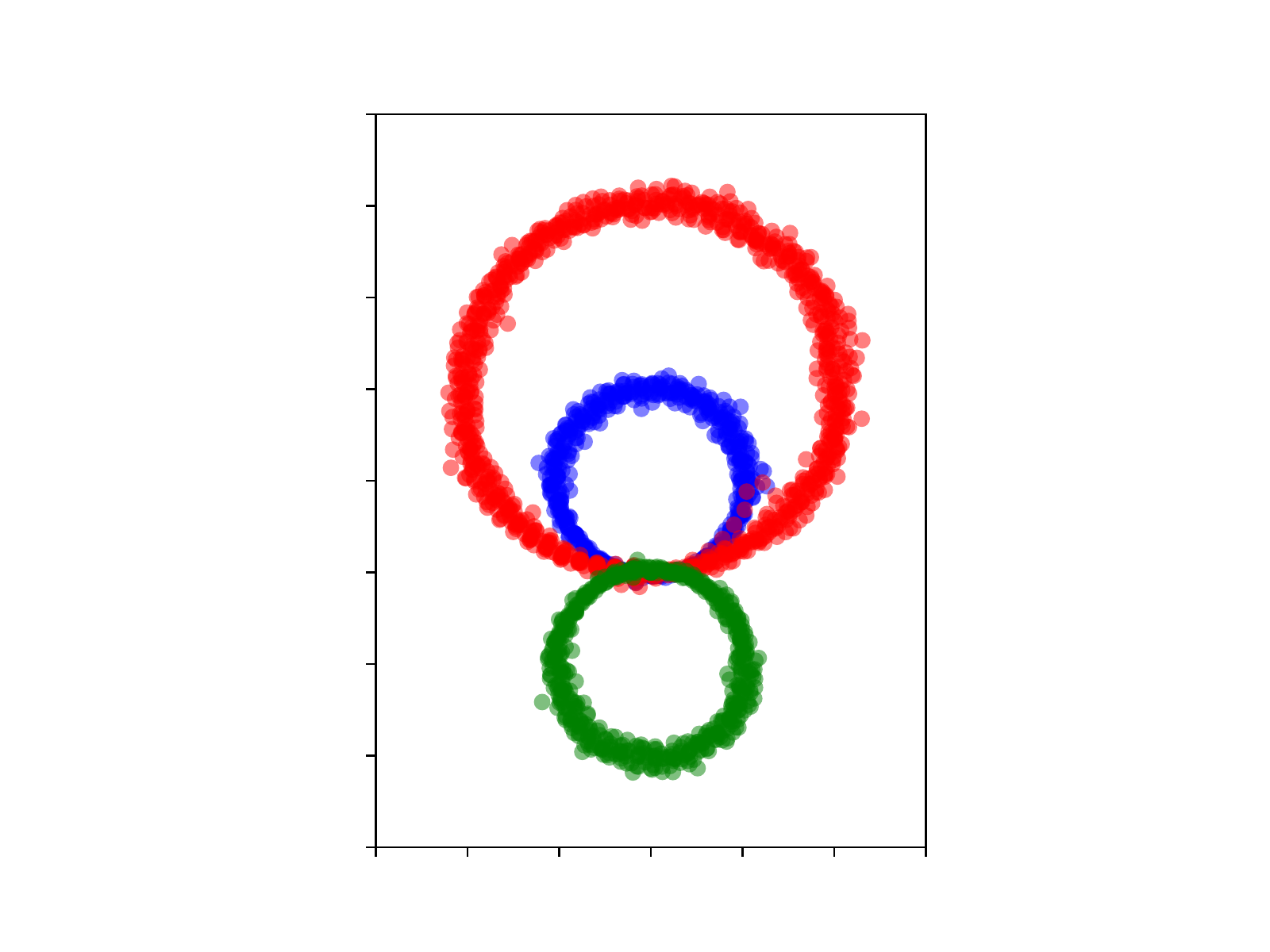}
\caption{Expert}
\end{subfigure}
~
\begin{subfigure}{0.23\textwidth}
\includegraphics[width=\textwidth]{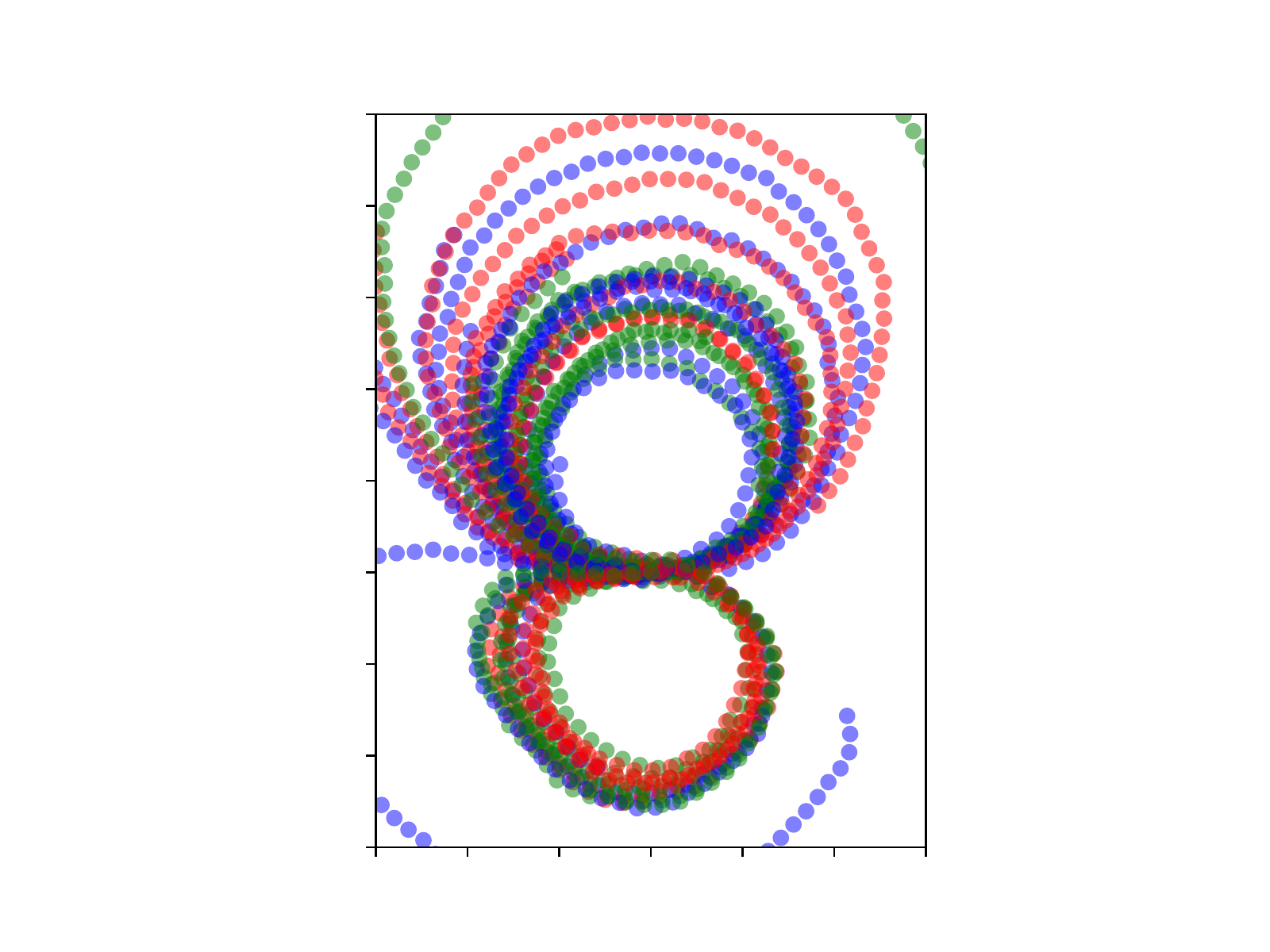}
\caption{Behavior cloning}
\end{subfigure}
~
\begin{subfigure}{0.23\textwidth}
\includegraphics[width=\textwidth]{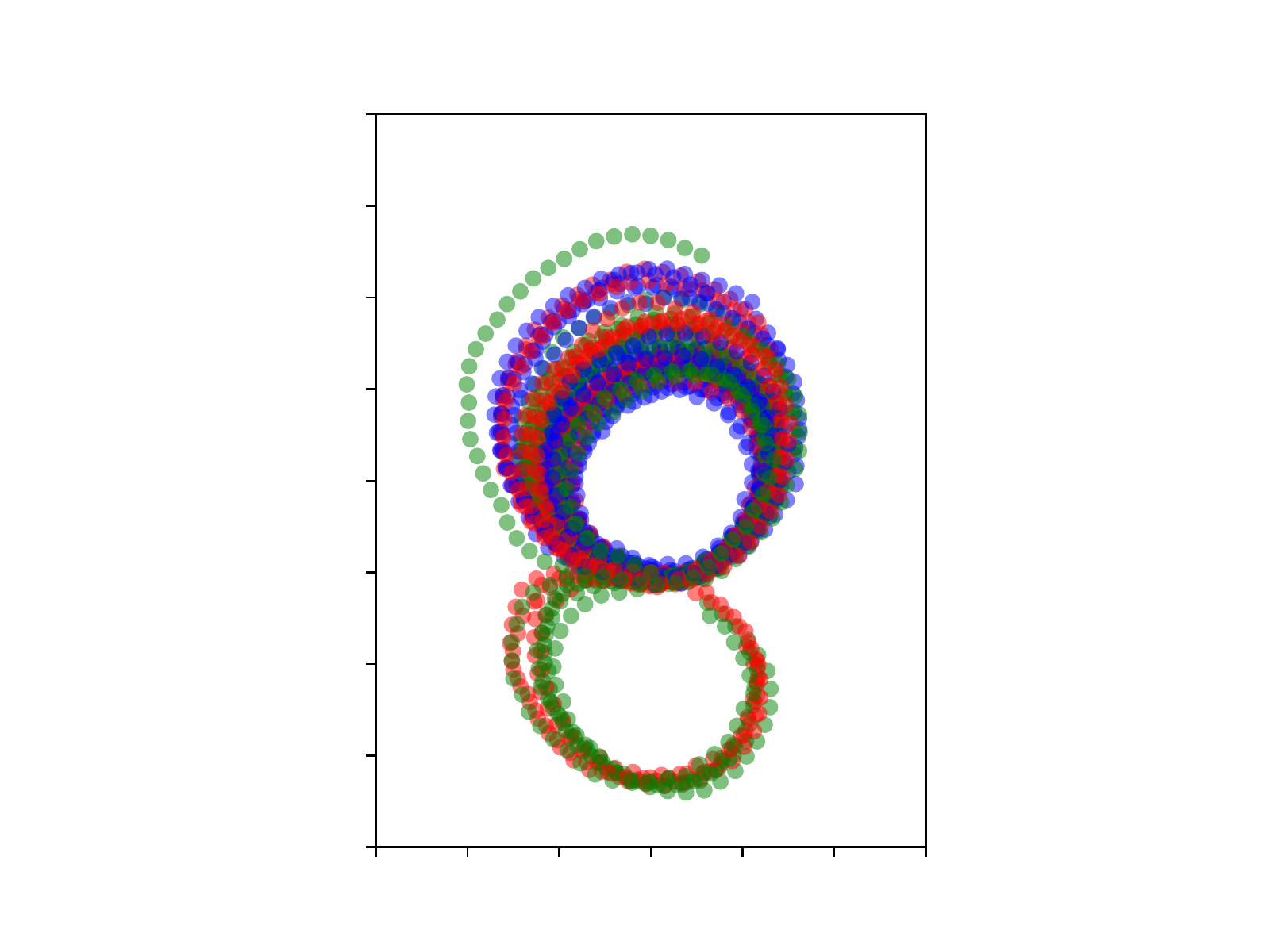}
\caption{GAIL}
\end{subfigure}
~
\begin{subfigure}{0.23\textwidth}
\includegraphics[width=\textwidth]{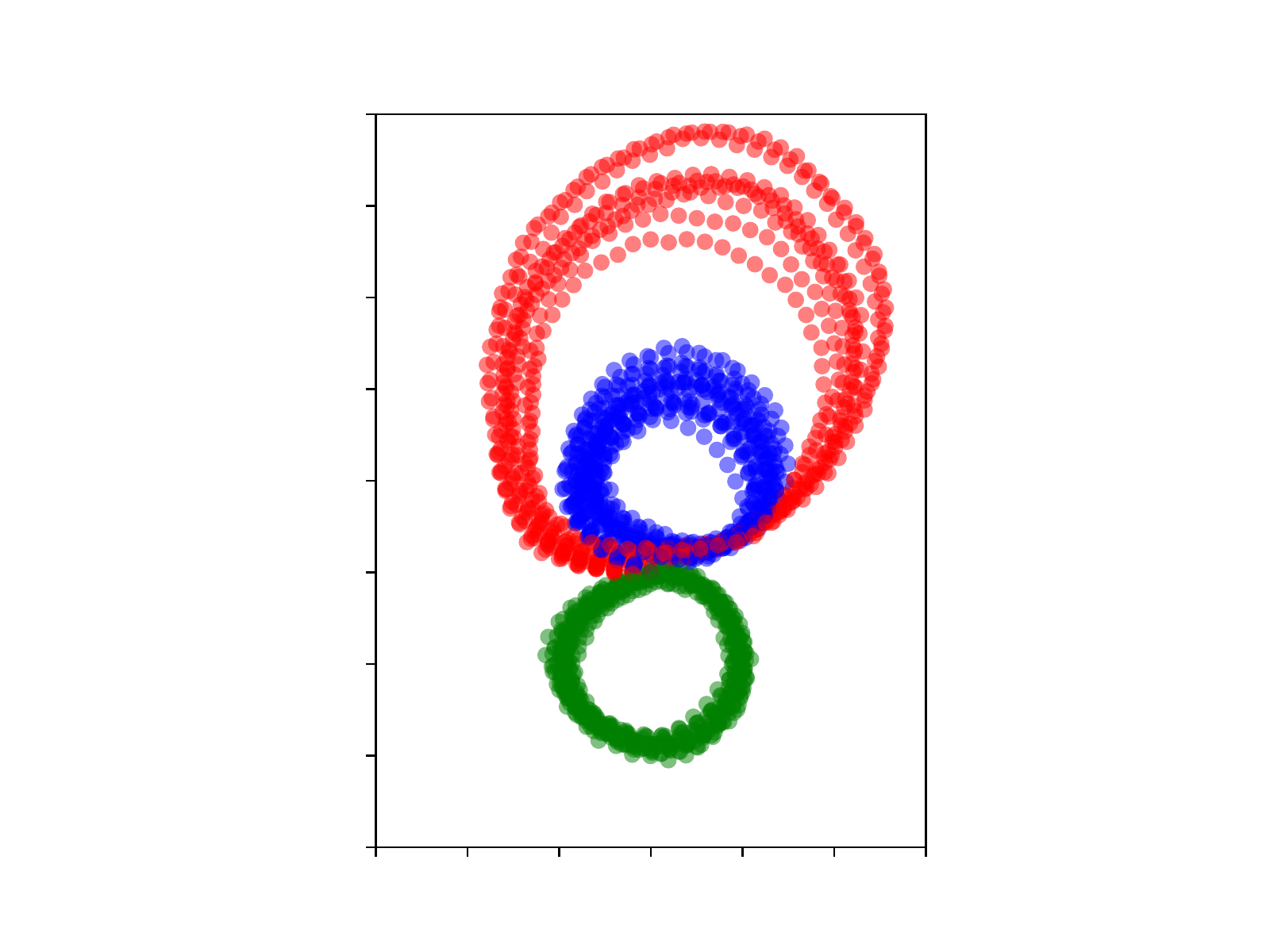}
\caption{Ours}
\end{subfigure}
\caption{{\bf Learned trajectories in the synthetic 2D plane environment.} Each color denotes one specific latent code. Behavior cloning deviates from the expert demonstrations due to compounding errors. %as errors within each step are accumulating along the trajectory.\se{what does this refer to?}. 
GAIL does produce circular trajectories but fails to capture the latent structure for it assumes that the demonstrations are generated from a single expert, and tries to learn an average policy. %\se{explain}. 
Our method (InfoGAIL) successfully distinguishes expert behaviors and imitates each mode accordingly (colors are ordered in accordance to the expert for visualization purposes, but are not identifiable).% \se{shouldnt everything be up to permutations?}
}
\label{fig:toy}
\end{figure}

\subsection{Learning to Distinguish Trajectories}
\label{sec:toy}
We demonstrate the effectiveness of InfoGAIL on a synthetic example. The environment is a 2D plane where the agent can move around freely at a constant velocity by selecting its direction $\vect{p}_t$ at (discrete) time $t$. For the agent, the observations at time $t$ are positions from $t-4$ to $t$. The (unlabeled) expert demonstrations contain three distinct modes, each generated with a stochastic expert policy that produces a circle-like trajectory (see Figure \ref{fig:toy}, panel a). %aims to stay on the circle (\se{be more precise}). 
The objective is to distinguish these three distinct modes and imitate the corresponding expert behavior. We consider three methods: behavior cloning, GAIL and InfoGAIL (details included in Appendix \ref{sec:toy-app}). In particular, for all the experiments we assume the same architecture and that the latent code is a one-hot encoded vector with 3 dimensions and a uniform prior; only InfoGAIL regularizes the latent code. Figure \ref{fig:toy} shows that the introduction of latent variables allows InfoGAIL to distinguish the three types of behavior and imitate each behavior successfully; the other two methods, however, fail to distinguish distinct modes. BC suffers from the compounding error problem and the learned policy tends to deviate from the expert trajectories; GAIL does learn to generate circular trajectories but it fails to separate different modes due to the lack of a mechanism that can explicitly account for the underlying structure. % \se{how do they fail?}

In the rest of Section \ref{sec:experiments}, we show how InfoGAIL can infer the latent structure of human decision-making in a driving domain. In particular, our agent only relies on visual inputs to sense the environment. 

\subsection{Utilizing Raw Visual Inputs via Transfer Learning}

%\se{this transfer learning stuff is very disconnected}

% Visual data is often inexpensive to obtain, highly informative, and heavily relied upon by people when performing tasks. Although our approach is general, we will focus on scenarios where the states $s$ are represented by images. Despite recent successes in visual perception using convolutional neural networks (CNNs)~\citep{krizhevsky2012imagenet}, this is a very challenging scenario as training a modern neural network is expensive and typically requires a large amount of training data. \js{might need to modify / remove above this}

The high dimensional nature of visual inputs poses a significant challenges to learning a policy.
Intuitively, the policy will have to simultaneously learn how to identify meaningful visual features, and how to leverage them to achieve the desired behavior using only a small number of expert demonstrations. Therefore, methods to mitigate the high sample complexity of the problem are crucial to success in this domain.

In this paper, we take a transfer learning approach. Features extracted using a CNN pre-trained on ImageNet contain high-level information about the input images, which can be adapted to new vision tasks via transfer learning~\citep{yosinski2014transferable}. However, it is not yet clear whether these relatively high-level features can be directly applied to tasks where perception and action are tightly interconnected; we demonstrate that this is possible through our experiments.  We perform transfer learning by exploiting features from a pre-trained neural network that effectively convert raw images into relatively high-level information~\citep{sharif2014cnn}. 
In particular, we use a Deep Residual Network~\citep{he2016deep} pre-trained on the ImageNet classification task \citep{russakovsky2015imagenet} to obtain the visual features used as inputs for the policy network.

\subsection{Network Structure}
Our policy accepts certain auxiliary information as internal input to serve as a short-term memory. 
This auxiliary information can be accessed along with the raw visual inputs.
In our experiments, the auxiliary information for the policy at time $t$ consists of the following: 1) \textbf{velocity} at time $t$, which is a three dimensional vector; 2) \textbf{actions} at time $t-1$ and $t-2$, which are both three dimensional vectors; 3) \textbf{damage} of the car, which is a real value. The auxiliary input has 10 dimensions in total.

For the policy network, input visual features are passed through two convolutional layers, and then combined with the auxiliary information vector and (in the case of InfoGAIL) the latent code $c$. 
%The exact architecture for $\pi_\theta$ is in Figure \ref{fig:policy}. 
%Moreover, merging latent codes at the higher levels would have less effect over the actions, since the visual features have much larger dimensions.
We parameterize the baseline as a network with the same architecture except for the final layer, which is just a scalar output that indicates the expected accumulated future rewards.

The discriminator $D_\omega$ accepts three elements as input: the input image,
%a resized image with lower resolution, 
the auxiliary information, and the current action. The output is a score for the WGAN training objective, which is supposed to be lower for expert state-action pairs, and higher for generated ones. 
%Details for the architecture of $D_\omega$ are shown in Figure \ref{fig:discriminator}. 
The posterior approximation network $Q_\psi$ adopts the same architecture as the discriminator, except that the output is a softmax over the discrete latent variables or a factored Gaussian over continuous latent variables. We include details of our architecture in Appendix \ref{sec:arch}.

\subsection{Interpretable Imitation Learning from Visual Demonstrations}
In this experiment, we consider two subsets of human driving behaviors: \textit{\textbf{turn}}, where the expert takes a turn using either the inside lane or the outside lane; and \textit{\textbf{pass}}, where the expert passes another vehicle from either the left or the right. In both cases, the expert policy has two significant modes.
Our goal is to have InfoGAIL capture these two separate modes from expert demonstrations in an unsupervised way.

We use a discrete latent code, which is a one-hot encoded vector with two possible states. For both settings, there are 80 expert trajectories in total, with 100 frames in each trajectory; our prior for the latent code is a uniform discrete distribution over the two states.
%We denote this setting as the short-term experiment \stefano{short term experiment sounds odd}.
%This is analogous to how humans learn to drive in that proficient drivers in the United States do not have to be entirely retrained on how to drive in the United Kingdoms, even though the two countries drive on different sides of the road.
The performance of a learned policy is quantified with two metrics: the \textit{average distance} is determined by the distance traveled by the agent before a collision (and is bounded by the length of the simulation horizon), 
%which is defined to be the probability that the agent manages to drive without collision, 
and \textit{accuracy} is defined as the classification accuracy of the expert state-action pairs according to the latent code inferred with $Q_\psi$. We add constant reward at every time step as reward augmentation, which is used to encourage the car to "stay alive" as long as possible and can be regarded as another way of reducing collision and off-lane driving (as these will lead to the termination of that episode).

\begin{figure*}[t!]
  \centering
  \includegraphics[width=0.95\textwidth]{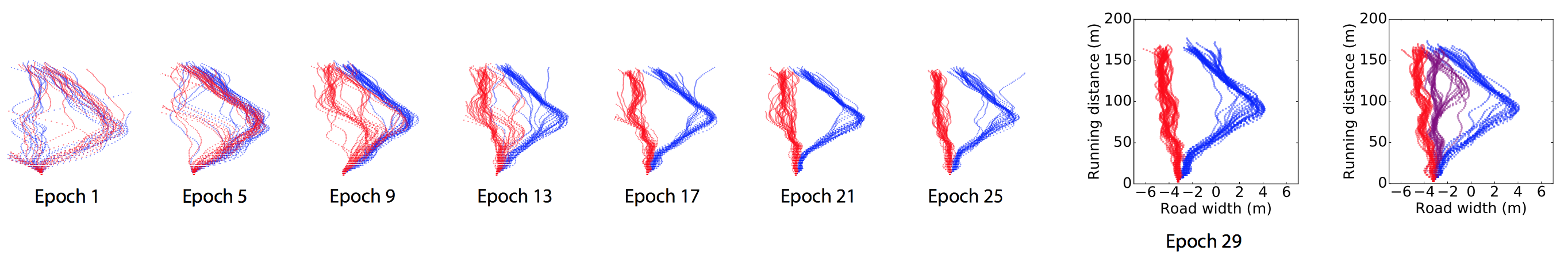}
  \caption{{\bf Visualizing the training process of \textit{turn}.} Here we show the trajectories of InfoGAIL at different stages of training. Blue and red indicate policies under different latent codes, which correspond to ``turning from inner lane'' and ``turning from outer lane'' respectively. 
  %The axes are the traveling distance and horizontal positions with respect to the road. Cutting a corner requires moving from one side to the other, and then moving back. 
  %Hence, blue indicates turning from the inner lane, and red indicates turning from the outer lane. 
  The rightmost figure shows the trajectories under latent codes $[1, 0]$ (red), $[0, 1]$ (blue), and $[0.5, 0.5]$ (purple), which suggests that, to some extent, our method is able to generalize to cases previously unseen in the training data.
  %\yunzhu{Note that here the axises are with respected to the road (traveling distance \& position in the road).}%Variance across different epochs is due to the limited number of rollouts, which are expensive to obtain with the TORCS simulator. 
  }
  \label{fig:vis_case_0}
\end{figure*}

\begin{figure*}[t!]
  \centering
  \includegraphics[width=\textwidth]{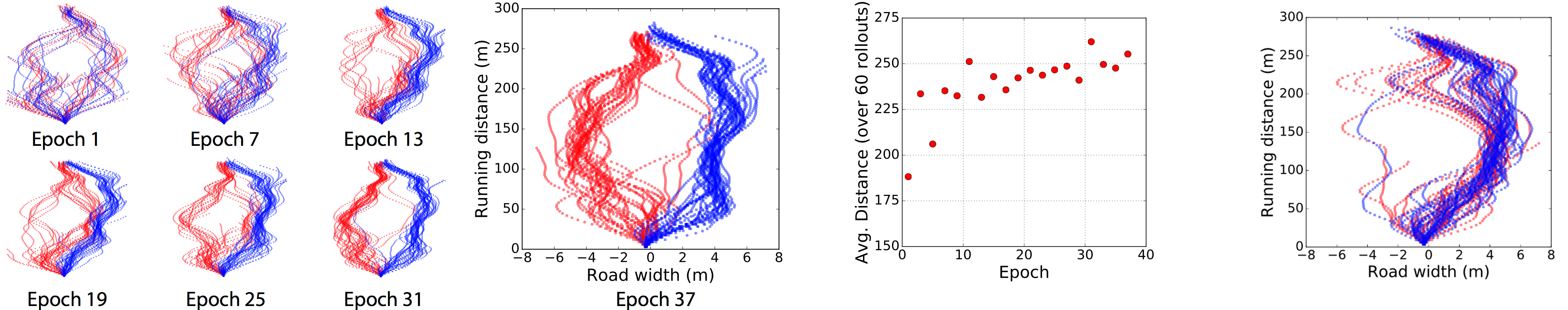}
  \caption{ \textbf{Experimental results for \textit{pass}}. \textbf{Left}: Trajectories of InfoGAIL at different stages of training (epoch 1 to 37). Blue and red indicate policies using different latent code values, which correspond to passing from right or left. \textbf{Middle}: Traveled distance denotes the absolute distance from the start position, % and is long since the other car is also moving quickly. 
 averaged over 60 rollouts of the InfoGAIL policy trained at different epochs. \textbf{Right}: Trajectories of \textit{\textbf{pass}} produced by an agent trained on the original GAIL objective. Compared to InfoGAIL, GAIL fails to distinguish between different modes.
  %Variance across different epochs is due to the limited number of rollouts, which are expensive to obtain with the TORCS simulator. 
  }
  \label{fig:vis_case_1}
\end{figure*}

The average distance and sampled trajectories at different stages of training are shown in Figures \ref{fig:vis_case_0} and \ref{fig:vis_case_1} for \textit{\textbf{turn}} and \textbf{\textit{\textbf{pass}}} respectively. During the initial stages of training, the model does not distinguish the two modes and has a high chance of colliding and driving off-lane, due to the limitations of behavior cloning (which we used to initialize the policy). As training progresses, trajectories provided by the learned policy begin to diverge. Towards the end of training, the two types of trajectories are clearly distinguishable, with only a few exceptions. In \textit{\textbf{turn}}, $[0, 1]$ corresponds to using the inside lane, while $[1, 0]$ corresponds to the outside lane. In \textit{\textbf{pass}}, the two kinds of latent codes correspond to passing from right and left respectively. Meanwhile, the average distance of the rollouts steadily increases with more training.

%In Figure \ref{fig:infogail}, we show the visual input for the policy at certain time steps, during which the agent observes itself passing a corner or surpassing another vehicle from the side determined by the latent code. This behavior is visually similar to the corresponding expert behavior.

Learning the two modes separately requires accurate inference of the latent code.
To examine the accuracy of posterior inference, we select state-action pairs from the expert trajectories (where the state is represented as a concatenation of raw image and auxiliary variables) and obtain the corresponding latent code through $Q_\psi(c|s, a)$; see Table~\ref{table:inference}. Although we did not explicitly provide any label, our model is able to correctly distinguish over $81\%$ of the state-action pairs in \textbf{\textit{pass}} (and almost all the pairs in \textbf{\textit{turn}}, confirming the clear separation between generated trajectories with different latent codes in Figure \ref{fig:vis_case_0}).
%\se{what can be seen in the figure?}
%The results for trajectories generated by $\pi_\theta$ are shown in Figure \ref{fig:vis_train}. During the initial stages of training, the model has not developed clear concept over different paths, and has high chances of failing, namely driving outside the track or colliding with the other vehicle. However, simply after 20 epochs of training, the policy has learned to distinguish surpassing from the left or the right, and is able to perform each policy with increasingly high success rates. We also show the visual input for the policy at certain time steps in Figure \ref{fig:infogail}, where the agent observes itself surpassing another vehicle from the side determined by the latent code. \td{more description of this; it seems very significant }
 
% \begin{figure*}
%   \centering
%   \includegraphics[width=0.8\textwidth]{infogail.pdf}
%   \caption{
%   Visual inputs for the InfoGAIL policies in \textit{\textbf{pass}} and \textit{\textbf{turn}} experiments. The latent code controls the behavior of the policies. %Videos are included in the supplementary material.
%   }
%   \label{fig:infogail}
%  \end{figure*}

For comparison, we also visualize the trajectories of \textit{\textbf{pass}} for the original GAIL objective in Figure \ref{fig:vis_case_1}, where there is no mutual information regularization. GAIL learns the expert trajectories as a whole, and cannot distinguish the two modes in the expert policy. 

Interestingly, instead of learning two separate trajectories, GAIL tries to fit the left trajectory by swinging the car suddenly to the left \textit{after} it has surpassed the other car from the right. We believe this reflects a limitation in the discriminators. Since $D_\omega(s, a)$ only requires state-action pairs as input, the policy is only required to match most of the state-action pairs; matching each rollout in a whole with expert trajectories is not necessary. InfoGAIL with discrete latent codes can alleviate this problem by forcing the model to learn separate trajectories.
%, but we believe this is an open problem for GAIL in cases where the expert policy is complex and has multiple modes.

%that only require state-action pairs as input, since $\pi_\theta$ can try to fit the expert state-action pairs by making sudden movements. This would fit the remaining state-actions pairs which would fool the discriminator, at the cost of a short, unreasonable action; although sudden movements makes no sense trajectory-wise, they will increase the overall score from a discriminator that only distinguishes state-action pairs. 

%  \begin{figure}
%   \centering
%   \includegraphics[width=0.3\textwidth]{imgs/gail-rollouts.png}
%   \caption{{\bf Visualizing GAIL rollouts.} Blue and red indicates policies under different latent codes. Changing the latent code has no influence over the resulting policy. \yunzhu{Need to be merged in Figure~\ref{fig:vis_case_1}.} }
%   \label{fig:gail_rollouts}
%  \end{figure}

\begin{table}
\begin{minipage}[t]{0.45\textwidth}
\caption{Classification accuracies for \textit{pass}.
%unsupervised and supervised methods for \textit{pass}. %Unsupervised methods are tested over all the expert trajectories; supervised methods are trained over $70\%$ of the trajectories and tested over the remaining $30\%$.
}
\label{table:inference}
\vskip 0.15in
\begin{center}
\begin{small}
%\begin{sc}
% \begin{tabular}{ll|ll}
% \hline
% Unsupervised & & Supervised & \\
% \hline
% K-means     & 55.4\%	    & SVM & 85.8\% \\
% PCA         & 61.7\%        & CNN & 90.8\% \\
% \textbf{\textsc{InfoGAIL (Ours)}}  & {\bf 81.9\%}  & & \\
% \hline
% \end{tabular}
%\end{sc}
\begin{tabular}{lc}
\toprule
Method & Accuracy \\\midrule
Chance		& 50\% \\
K-means     & 55.4\% \\
PCA         & 61.7\% \\
\textbf{InfoGAIL (Ours)}  & {\bf 81.9\%} \\\midrule
SVM & 85.8\% \\
\textbf{CNN} & \textbf{90.8}\% \\
\bottomrule
\end{tabular}
\end{small}
\end{center}
\vskip -0.1in
\end{minipage}
~
\begin{minipage}[t]{0.48\textwidth}
\caption{Average rollout distances.}
\label{table:ablation}
%\vskip 0.15in
\begin{center}
\begin{small}
%\begin{sc}
\begin{tabular}{lc}
\toprule
Method & Avg. rollout distance \\
\midrule
Behavior Cloning			& 701.83 \\
GAIL & 914.45 \\
InfoGAIL $\setminus$ RB		& 1031.13 \\
InfoGAIL $\setminus$ RA		& 1123.89 \\
InfoGAIL $\setminus$ WGAN   & 1177.72 \\
\textbf{InfoGAIL (Ours)}		& {\bf 1226.68} \\
Human						& 1203.51 \\
\bottomrule
\end{tabular}
%\end{sc}
\end{small}
\end{center}
\vskip -0.1in
\end{minipage}
\end{table}

\subsection{Ablation Experiments}
%\td{subsection that talks about controlling the optimization factors}
We conduct a series of ablation experiments to demonstrate that our proposed improved optimization techniques in Section \ref{sec:rewardaugmentation} and \ref{sec:optimization} are indeed crucial for learning an effective policy. %The experiments consider a long-term setting: 
Our policy drives a car on the race track along with other cars, whereas the human expert provides 20 trajectories with 500 frames each by trying to drive as fast as possible without collision. Reward augmentation is performed by adding a reward that encourages the car to drive faster. The performance of the policy is determined by the average distance. Here a longer average rollout distance indicates a better policy.

In our ablation experiments, we selectively remove some of the improved optimization methods from Section \ref{sec:rewardaugmentation} and \ref{sec:optimization} (we do not use any latent code in these experiments). \textbf{InfoGAIL(Ours)} includes all the optimization techniques; \textbf{GAIL} excludes all the techniques; \textbf{InfoGAIL$\setminus$WGAN} switches the WGAN objective with the GAN objective; \textbf{InfoGAIL$\setminus$RA} removes reward augmentation; \textbf{InfoGAIL$\setminus$RB} removes the replay buffer and only samples from the most recent rollouts; \textbf{Behavior Cloning} is the behavior cloning method and \textbf{Human} is the expert policy.
Table \ref{table:ablation} shows the average rollout distances of different policies. Our method is able to outperform the expert with the help of reward augmentation; policies without reward augmentation or WGANs perform slightly worse than the expert; removing the replay buffer causes the performance to deteriorate significantly due to increased variance in gradient estimation.

% \begin{figure*}
%   \centering
%   \includegraphics[width=0.8\textwidth]{gail_pixel.pdf}
%   \caption{\textbf{Visual inputs for our policy in the ablation experiment.} During the rollout, our policy is able to pass several other cars. A video is included in the supplementary material.}
%   \label{fig:gail_pixel}
%  \end{figure*}

%\td{need some criteria to measure performance}
%\td{controlling the factors: image feature, wgan, reward augmentation, smoothing}

\section{Related work}

%\stefano{cite the original pomerleanu driving stuff}

There are two major paradigms for vision-based driving systems ~\citep{chen2015deepdriving}. \textit{Mediated perception} is a two-step approach that first obtains scene information and then makes a driving decision ~\citep{aly2008real,lenz2011sparse,kitani2012activity}; \textit{behavior reflex}, on the other hand, adopts a direct approach by mapping visual inputs to driving actions ~\citep{pomerleau1989alvinn,pomerleau1991efficient}. Many of the current autonomous driving methods rely on the two-step approach, which requires hand-crafting features such as the detection of lane markings and cars ~\citep{geiger2013vision,chen2015deepdriving}. Our approach, on the other hand, attempts to learn these features directly from vision to actions. While mediated perception approaches are currently more prevalent, we believe that end-to-end learning methods are more scalable and may lead to better performance in the long run.

\cite{bojarski2016end} introduce an end-to-end imitation learning framework that learns to drive entirely from visual information, and test their approach on real-world scenarios. However, their method uses behavior cloning by performing supervised learning over the state-action pairs, which is well-known to generalize poorly 
%due to the potential of compounding errors. To avoid the car from slowly drifting away, they augment the visual information by introducing three cameras in a row,
%during; if the car deviates from its lane, the cameras on the left and right will capture this information, in which case the car will be trained to steer back to the middle of the lane. This approach 
%which requires additional visual information 
%and have difficulties generalizing 
to more sophisticated tasks, such as changing lanes or passing vehicles. With the use of GAIL, our method can learn to perform these sophisticated operations easily.
\cite{zhang2016query} performs end-to-end visual imitation learning in TORCS through DAgger \cite{ross2011reduction}, querying the reference policies during training, which in many cases is difficult.

%; we also require visual inputs from only one camera, which are much easier to collect in real world scenarios. \stefano{try to summarize this a bit}

Most imitation learning methods for end-to-end driving rely heavily on LIDAR-like inputs to obtain precise distance measurements~\citep{ho2016model,kuefler2017imitating}. These inputs are not usually available to humans during driving. In particular, \cite{kuefler2017imitating} applies GAIL to the task of modeling human driving behavior on highways. %Their policy is modeled using a recurrent neural network, which is supposed to maintain sufficient statistics of the past observations.
In contrast, our policy requires only \textit{raw visual information} as external input, which in practice is all the information humans need in order to drive. %Additionally, the success in our policy architecture demonstrate that autonomous vehicles might not require a sophisticated memory to simply drive forward.

\cite{sermanet2016unsupervised} and \cite{finn2016guided} have also introduced a pre-trained deep neural network to achieve better performance in imitation learning with relatively few demonstrations. Specifically, they introduce a pre-trained model to learn dense, incremental reward functions that are suitable for performing downstream reinforcement learning tasks, such as real-world robotic experiments. This is different from our approach, in that transfer learning is performed over the critic instead of the policy. %Since pre-trained neural networks have displayed the potential to learn more sophisticated reward functions, %that match the reinforcement learning problem
It would be interesting to combine that reward with our approach through reward augmentation. %\td{but this reward also comes from the expert trajectory, so is it a legitimate claim?}

% Inverse Reinforcement learning \cite{abbeel2004apprenticeship}
% Trust region policy optimization \cite{schulman2015trust}
% Generative adversarial networks \cite{goodfellow2014generative}
% Generative adversarial imitation learning \cite{ho2016generative}
% Deep deterministic policy gradient \cite{lillicrap2015continuous}
% Generalized advantage estimation \cite{schulman2015high}
% Third person imitation learning \cite{Stadie2017third}
% Imitating driving behavior with generative adversarial networks \cite{kuefler2017imitating}
% InfoGAN \cite{chen2016infogan}
% DeepDriving \cite{chen2015deepdriving}
% Model-free imitation learning with policy optimization \cite{ho2016model}
% Behavior cloning \cite{pomerleau1991efficient}
% Compounding error caused by covariance shift \cite{ross2010efficient,ross2011reduction}
% Torcs \cite{wymann2000torcs}
% Resnet \cite{he2016deep}
% ImageNet \cite{deng2009imagenet}
% Convolutional neural networks \cite{krizhevsky2012imagenet}
% Wasserstein GAN \cite{arjovsky2017wasserstein}

\section{Conclusion}
In this paper, we present a method to imitate complex behaviors while identifying salient latent factors of variation in the demonstrations. Discovering these latent factors does not require direct supervision beyond expert demonstrations, and the whole process can be trained directly with standard policy optimization algorithms. We also introduce several techniques to successfully perform imitation learning using visual inputs, including transfer learning and reward augmentation. Our experimental results in the TORCS simulator show that our methods can automatically distinguish certain behaviors in human driving, while learning a policy that can imitate and even outperform the human experts using visual information as the sole external input. We hope that our work can further inspire end-to-end learning approaches to autonomous driving under more realistic scenarios.

%Although we demonstrate success using discrete latent codes, our method cannot be generalized to continuous latent codes directly, since the discriminator fails to take trajectory-level information into account. One interesting future work would be extending GAIL to imitate trajectories with temporal information, which would also extend InfoGAIL to continuous latent codes.

%Another compelling direction for future work is to explore how current mediated perception approaches could be combined with direct imitation learning through reward augmentation. Scene information could provide us with more powerful reinforcement signals, which are crucial to training machines that are better at driving than humans.
\vspace{-.5em}
\subsection*{Acknowledgements}
We thank Shengjia Zhao and Neal Jean for their assistance and advice. %We also thank the anonymous reviewers for their helpful comments. 
Toyota Research Institute (TRI) provided funds to assist the authors with their research but this article solely reflects the opinions and conclusions of its authors and not TRI or any other Toyota entity. This research was also supported by Intel Corporation, FLI and NSF grants 1651565, 1522054, 1733686. 

\bibliographystyle{ieeetr}
\bibliography{nips_2017}

\begin{thebibliography}{10}

\bibitem{levine2013guided}
S.~Levine and V.~Koltun, ``Guided policy search.,'' in {\em ICML (3)},
  pp.~1--9, 2013.

\bibitem{schulman2015trust}
J.~Schulman, S.~Levine, P.~Abbeel, M.~I. Jordan, and P.~Moritz, ``Trust region
  policy optimization.,'' in {\em ICML}, pp.~1889--1897, 2015.

\bibitem{lillicrap2015continuous}
T.~P. Lillicrap, J.~J. Hunt, A.~Pritzel, N.~Heess, T.~Erez, Y.~Tassa,
  D.~Silver, and D.~Wierstra, ``Continuous control with deep reinforcement
  learning,'' {\em arXiv preprint arXiv:1509.02971}, 2015.

\bibitem{schulman2015high}
J.~Schulman, P.~Moritz, S.~Levine, M.~Jordan, and P.~Abbeel, ``High-dimensional
  continuous control using generalized advantage estimation,'' {\em arXiv
  preprint arXiv:1506.02438}, 2015.

\bibitem{silver2016mastering}
D.~Silver, A.~Huang, C.~J. Maddison, A.~Guez, L.~Sifre, G.~Van Den~Driessche,
  J.~Schrittwieser, I.~Antonoglou, V.~Panneershelvam, M.~Lanctot, {\em et~al.},
  ``Mastering the game of go with deep neural networks and tree search,'' {\em
  Nature}, vol.~529, no.~7587, pp.~484--489, 2016.

\bibitem{tamar2016value}
A.~Tamar, S.~Levine, P.~Abbeel, Y.~WU, and G.~Thomas, ``Value iteration
  networks,'' in {\em Advances in Neural Information Processing Systems},
  pp.~2146--2154, 2016.

\bibitem{ziebart2008maximum}
B.~D. Ziebart, A.~L. Maas, J.~A. Bagnell, and A.~K. Dey, ``Maximum entropy
  inverse reinforcement learning.,'' in {\em AAAI}, vol.~8, pp.~1433--1438,
  Chicago, IL, USA, 2008.

\bibitem{englert2015inverse}
P.~Englert and M.~Toussaint, ``Inverse kkt--learning cost functions of
  manipulation tasks from demonstrations,'' in {\em Proceedings of the
  International Symposium of Robotics Research}, 2015.

\bibitem{finn2016guided}
C.~Finn, S.~Levine, and P.~Abbeel, ``Guided cost learning: Deep inverse optimal
  control via policy optimization,'' in {\em Proceedings of the 33rd
  International Conference on Machine Learning}, vol.~48, 2016.

\bibitem{Stadie2017third}
B.~Stadie, P.~Abbeel, and I.~Sutskever, ``Third person imitation learning,'' in
  {\em ICLR}, 2017.

\bibitem{ermon2015learning}
S.~Ermon, Y.~Xue, R.~Toth, B.~N. Dilkina, R.~Bernstein, T.~Damoulas, P.~Clark,
  S.~DeGloria, A.~Mude, C.~Barrett, {\em et~al.}, ``Learning large-scale
  dynamic discrete choice models of spatio-temporal preferences with
  application to migratory pastoralism in east africa.,'' in {\em AAAI},
  pp.~644--650, 2015.

\bibitem{ho2016generative}
J.~Ho and S.~Ermon, ``Generative adversarial imitation learning,'' in {\em
  Advances in Neural Information Processing Systems}, pp.~4565--4573, 2016.

\bibitem{goodfellow2014generative}
I.~Goodfellow, J.~Pouget-Abadie, M.~Mirza, B.~Xu, D.~Warde-Farley, S.~Ozair,
  A.~Courville, and Y.~Bengio, ``Generative adversarial nets,'' in {\em
  Advances in neural information processing systems}, pp.~2672--2680, 2014.

\bibitem{chen2016infogan}
X.~Chen, Y.~Duan, R.~Houthooft, J.~Schulman, I.~Sutskever, and P.~Abbeel,
  ``Infogan: Interpretable representation learning by information maximizing
  generative adversarial nets,'' in {\em Advances in Neural Information
  Processing Systems}, pp.~2172--2180, 2016.

\bibitem{wymann2000torcs}
B.~Wymann, E.~Espi{\'e}, C.~Guionneau, C.~Dimitrakakis, R.~Coulom, and
  A.~Sumner, ``Torcs, the open racing car simulator,'' {\em Software available
  at http://torcs. sourceforge. net}, 2000.

\bibitem{pomerleau1991efficient}
D.~A. Pomerleau, ``Efficient training of artificial neural networks for
  autonomous navigation,'' {\em Neural Computation}, vol.~3, no.~1, pp.~88--97,
  1991.

\bibitem{ross2010efficient}
S.~Ross and D.~Bagnell, ``Efficient reductions for imitation learning.,'' in
  {\em AISTATS}, pp.~3--5, 2010.

\bibitem{ross2011reduction}
S.~Ross, G.~J. Gordon, and D.~Bagnell, ``A reduction of imitation learning and
  structured prediction to no-regret online learning.,'' in {\em AISTATS},
  p.~6, 2011.

\bibitem{abbeel2004apprenticeship}
P.~Abbeel and A.~Y. Ng, ``Apprenticeship learning via inverse reinforcement
  learning,'' in {\em Proceedings of the twenty-first international conference
  on Machine learning}, p.~1, ACM, 2004.

\bibitem{syed2008apprenticeship}
U.~Syed, M.~Bowling, and R.~E. Schapire, ``Apprenticeship learning using linear
  programming,'' in {\em Proceedings of the 25th international conference on
  Machine learning}, pp.~1032--1039, ACM, 2008.

\bibitem{ho2016model}
J.~Ho, J.~K. Gupta, and S.~Ermon, ``Model-free imitation learning with policy
  optimization,'' in {\em Proceedings of the 33rd International Conference on
  Machine Learning}, 2016.

\bibitem{bloem2014infinite}
M.~Bloem and N.~Bambos, ``Infinite time horizon maximum causal entropy inverse
  reinforcement learning,'' in {\em Decision and Control (CDC), 2014 IEEE 53rd
  Annual Conference on}, pp.~4911--4916, IEEE, 2014.

\bibitem{kingma2014adam}
D.~Kingma and J.~Ba, ``Adam: A method for stochastic optimization,'' {\em arXiv
  preprint arXiv:1412.6980}, 2014.

\bibitem{salimans2016improved}
T.~Salimans, I.~Goodfellow, W.~Zaremba, V.~Cheung, A.~Radford, and X.~Chen,
  ``Improved techniques for training gans,'' in {\em Advances in Neural
  Information Processing Systems}, pp.~2234--2242, 2016.

\bibitem{arora2017generalization}
S.~Arora, R.~Ge, Y.~Liang, T.~Ma, and Y.~Zhang, ``Generalization and
  equilibrium in generative adversarial nets (gans),'' {\em arXiv preprint
  arXiv:1703.00573}, 2017.

\bibitem{arjovsky2017wasserstein}
M.~Arjovsky, S.~Chintala, and L.~Bottou, ``Wasserstein gan,'' {\em arXiv
  preprint arXiv:1701.07875}, 2017.

\bibitem{williams1992simple}
R.~J. Williams, ``Simple statistical gradient-following algorithms for
  connectionist reinforcement learning,'' {\em Machine learning}, vol.~8,
  no.~3-4, pp.~229--256, 1992.

\bibitem{mnih2015human}
V.~Mnih, K.~Kavukcuoglu, D.~Silver, A.~A. Rusu, J.~Veness, M.~G. Bellemare,
  A.~Graves, M.~Riedmiller, A.~K. Fidjeland, G.~Ostrovski, {\em et~al.},
  ``Human-level control through deep reinforcement learning,'' {\em Nature},
  vol.~518, no.~7540, pp.~529--533, 2015.

\bibitem{yosinski2014transferable}
J.~Yosinski, J.~Clune, Y.~Bengio, and H.~Lipson, ``How transferable are
  features in deep neural networks?,'' in {\em Advances in neural information
  processing systems}, pp.~3320--3328, 2014.

\bibitem{sharif2014cnn}
A.~Sharif~Razavian, H.~Azizpour, J.~Sullivan, and S.~Carlsson, ``Cnn features
  off-the-shelf: an astounding baseline for recognition,'' in {\em Proceedings
  of the IEEE Conference on Computer Vision and Pattern Recognition Workshops},
  pp.~806--813, 2014.

\bibitem{he2016deep}
K.~He, X.~Zhang, S.~Ren, and J.~Sun, ``Deep residual learning for image
  recognition,'' in {\em Proceedings of the IEEE Conference on Computer Vision
  and Pattern Recognition}, pp.~770--778, 2016.

\bibitem{russakovsky2015imagenet}
O.~Russakovsky, J.~Deng, H.~Su, J.~Krause, S.~Satheesh, S.~Ma, Z.~Huang,
  A.~Karpathy, A.~Khosla, M.~Bernstein, {\em et~al.}, ``Imagenet large scale
  visual recognition challenge,'' {\em International Journal of Computer
  Vision}, vol.~115, no.~3, pp.~211--252, 2015.

\bibitem{chen2015deepdriving}
C.~Chen, A.~Seff, A.~Kornhauser, and J.~Xiao, ``Deepdriving: Learning
  affordance for direct perception in autonomous driving,'' in {\em Proceedings
  of the IEEE International Conference on Computer Vision}, pp.~2722--2730,
  2015.

\bibitem{aly2008real}
M.~Aly, ``Real time detection of lane markers in urban streets,'' in {\em
  Intelligent Vehicles Symposium, 2008 IEEE}, pp.~7--12, IEEE, 2008.

\bibitem{lenz2011sparse}
P.~Lenz, J.~Ziegler, A.~Geiger, and M.~Roser, ``Sparse scene flow segmentation
  for moving object detection in urban environments,'' in {\em Intelligent
  Vehicles Symposium (IV), 2011 IEEE}, pp.~926--932, IEEE, 2011.

\bibitem{kitani2012activity}
K.~Kitani, B.~Ziebart, J.~Bagnell, and M.~Hebert, ``Activity forecasting,''
  {\em Computer Vision--ECCV 2012}, pp.~201--214, 2012.

\bibitem{pomerleau1989alvinn}
D.~A. Pomerleau, ``Alvinn, an autonomous land vehicle in a neural network,''
  tech. rep., Carnegie Mellon University, Computer Science Department, 1989.

\bibitem{geiger2013vision}
A.~Geiger, P.~Lenz, C.~Stiller, and R.~Urtasun, ``Vision meets robotics: The
  kitti dataset,'' {\em The International Journal of Robotics Research},
  vol.~32, no.~11, pp.~1231--1237, 2013.

\bibitem{bojarski2016end}
M.~Bojarski, D.~Del~Testa, D.~Dworakowski, B.~Firner, B.~Flepp, P.~Goyal, L.~D.
  Jackel, M.~Monfort, U.~Muller, J.~Zhang, {\em et~al.}, ``End to end learning
  for self-driving cars,'' {\em arXiv preprint arXiv:1604.07316}, 2016.

\bibitem{zhang2016query}
J.~Zhang and K.~Cho, ``Query-efficient imitation learning for end-to-end
  autonomous driving,'' {\em arXiv preprint arXiv:1605.06450}, 2016.

\bibitem{kuefler2017imitating}
A.~Kuefler, J.~Morton, T.~Wheeler, and M.~Kochenderfer, ``Imitating driver
  behavior with generative adversarial networks,'' {\em arXiv preprint
  arXiv:1701.06699}, 2017.

\bibitem{sermanet2016unsupervised}
P.~Sermanet, K.~Xu, and S.~Levine, ``Unsupervised perceptual rewards for
  imitation learning,'' {\em arXiv preprint arXiv:1612.06699}, 2016.

\end{thebibliography}

\newpage
\appendix
\section{Details for Experiment in Section \ref{sec:toy}}
\label{sec:toy-app}
We use \textit{conv} to denote a convolutional layer, and \textit{fc} to denote a fully connected layer.
For the 2D synthetic example, we initialize our InfoGAIL network from behavior cloning and did not use replay buffers for it is cheap to rollout in this setting. The observations are 10 dimensional while the actions are 2 dimensional. Similar to the TORCS experiments, we use Wasserstein GANs during training. The video showing the learning process of InfoGAIL is available in the supplemented video.

\begin{figure}[h]
\centering
\begin{subfigure}[t]{0.48\textwidth}
\centering
\includegraphics[width=\textwidth]{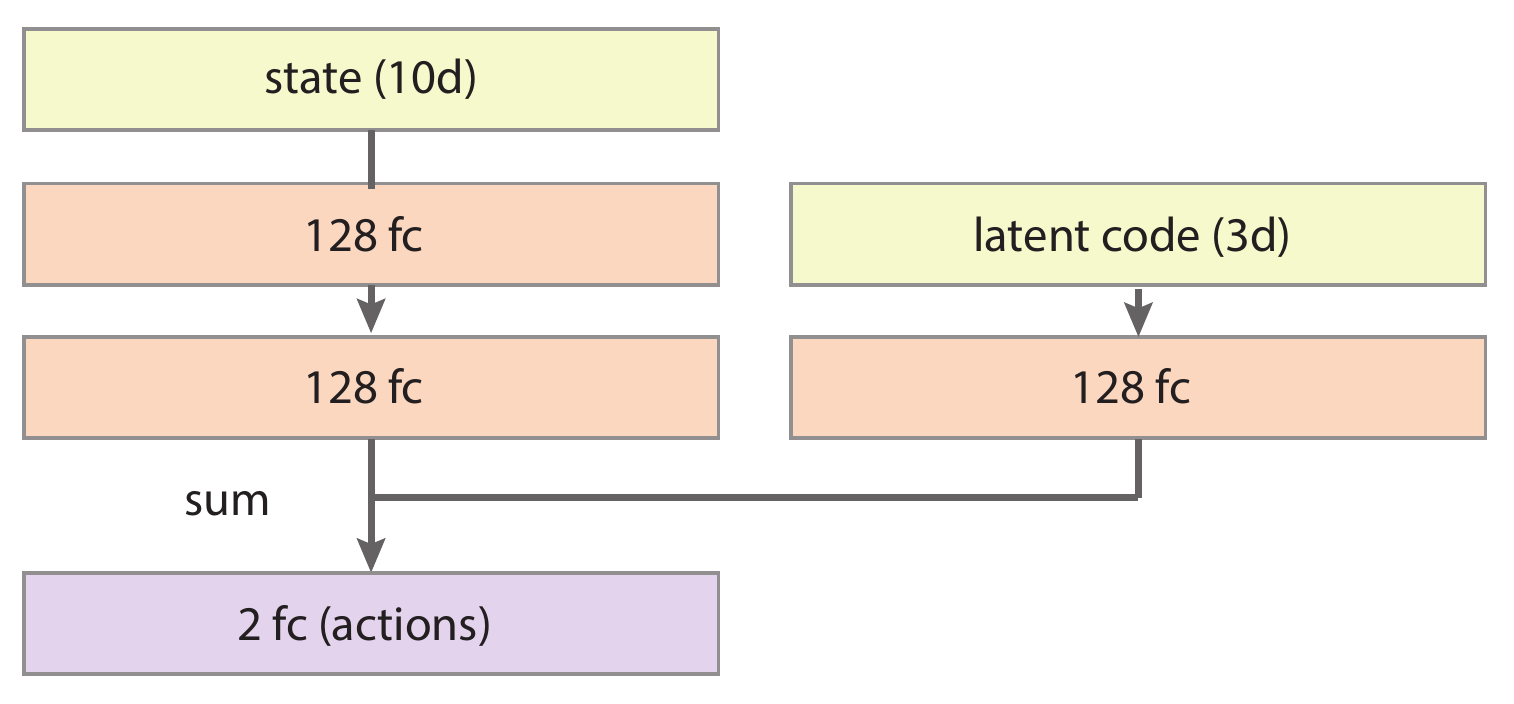}
\caption{Network architecture for the policy/generator $\pi_\theta$.} %\textit{conv} denotes a convolutional layer, and \textit{fc} denotes a fully connected layer.}

\label{fig:policy}
\end{subfigure}
~
\begin{subfigure}[t]{0.48\textwidth}
\centering
\includegraphics[width=0.5\textwidth]{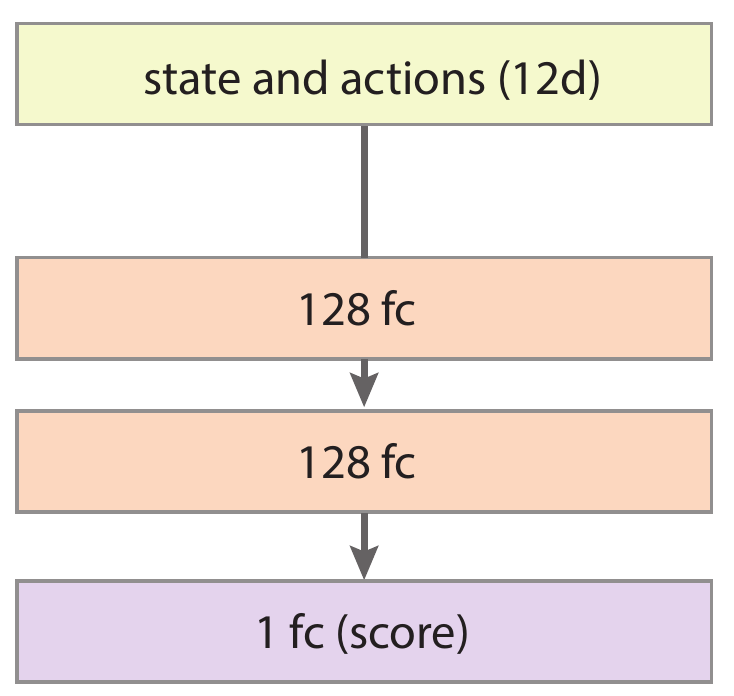}
\caption{Network architecture for the discriminator $D_\omega$.}
\label{fig:discriminator}
\end{subfigure}
\caption{Network architecture for the 2D synthetic environment.}
\end{figure}

\section{Architectural Details For TORCS Experiment}
\label{sec:arch}
We provide an input image with lower resolution to the discriminator mainly for improved training speed. Since inferring the high-level actions of the policy does not involve fine-grained details present in higher resolution inputs, we can improve training speed without suffering from performance loss.

\begin{figure}[h]
\centering
\begin{subfigure}[t]{0.48\textwidth}
\centering
\includegraphics[width=\textwidth]{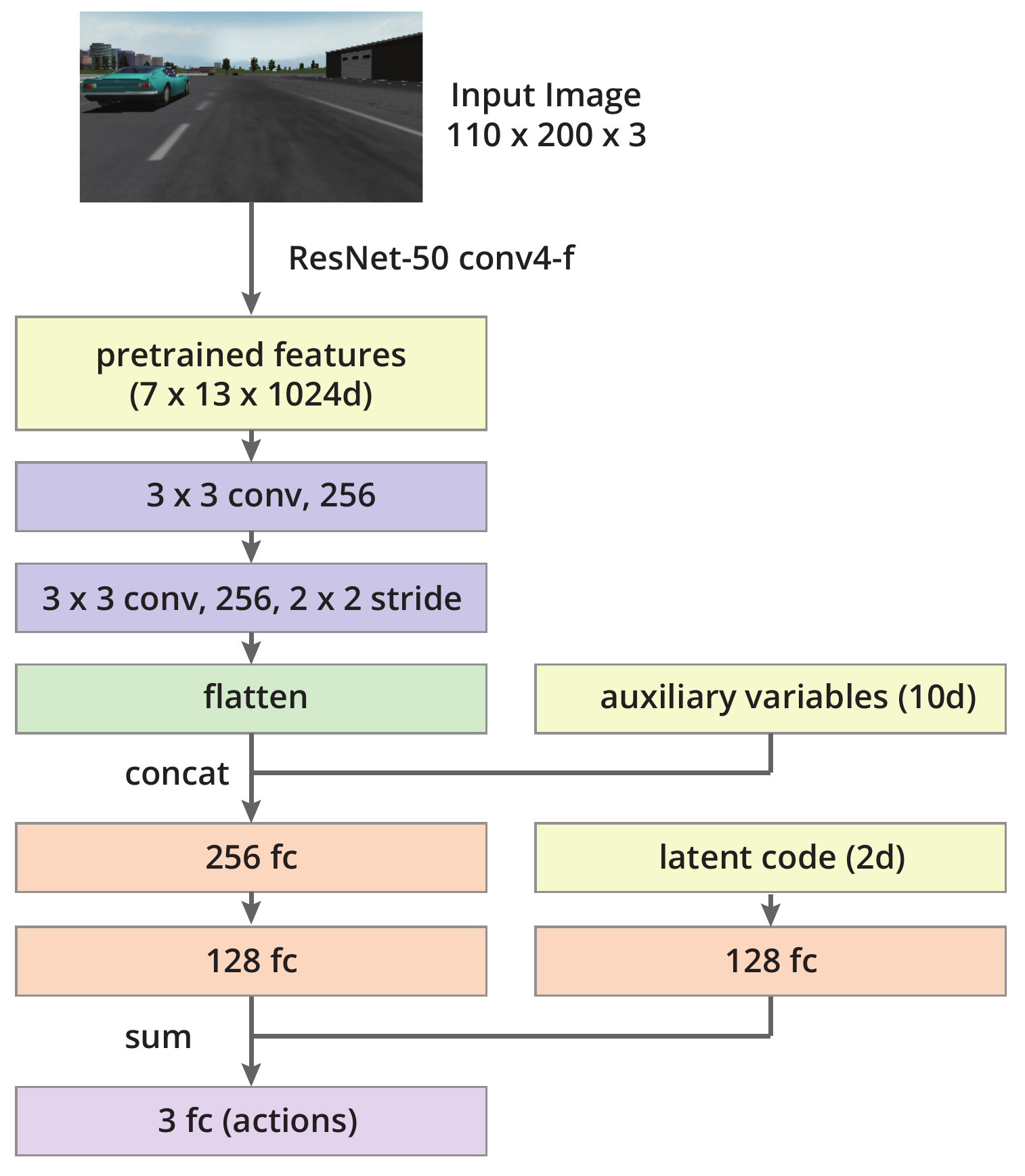}
\caption{Network architecture for the policy/generator $\pi_\theta$.} %\textit{conv} denotes a convolutional layer, and \textit{fc} denotes a fully connected layer.}

\label{fig:policy}
\end{subfigure}
~
\begin{subfigure}[t]{0.48\textwidth}
\centering
\includegraphics[width=\textwidth]{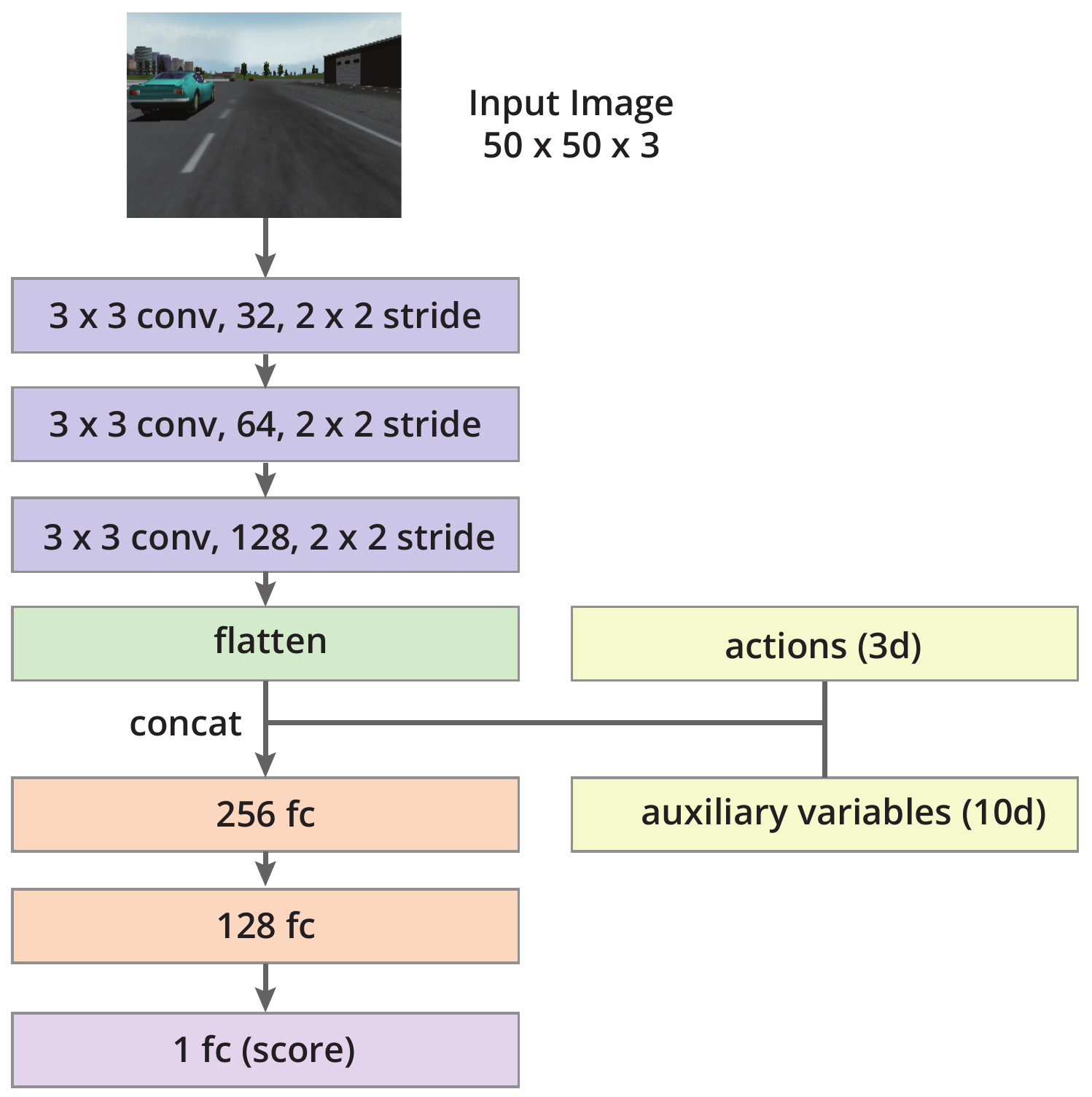}
\caption{Network architecture for the discriminator $D_\omega$.}
\label{fig:discriminator}
\end{subfigure}
\caption{Network architecture for the driving experiments.}
\end{figure}

\newpage
\section{Algorithm Outline for InfoGAIL}
\label{sec:infogail-algo}

An outline for the InfoGAIL algorithm with the discussed improvements is in Algorithm \ref{alg:infogail-ext}. 
\begin{algorithm}[h]
   \caption{InfoGAIL with extensions}
   \label{alg:infogail-ext}
\begin{algorithmic}
   \STATE {\bfseries Input:} Expert trajectories $\tau_E \sim \pi_E$; initial policy, discriminator and posterior parameters $\theta_0, \omega_0, \psi_0$; replay buffer $B = \varnothing$;
   \STATE \textbf{Output: } Learned policy $\pi_\theta$
   \FOR{$i=0, 1, 2, ... $}
   	   \STATE Sample a batch of latent codes: $c_i \sim P(c)$
       \STATE Sample trajectories: $\tau_i\sim\pi_{\theta_i}(c_i)$, with the latent code fixed during each rollout.
       \STATE Update the replay buffer: $B \gets B \cup \tau_i$.
       \STATE Sample $\chi_i \sim B$ and $\chi_E \sim \tau_E$ with same batch size.
%       \STATE Sample the same amount of state-action pairs $\chi_i$ from replay buffer, and $\chi_E$ from expert trajectories.
       \STATE Update $\omega_{i}$ to $\omega_{i+1}$ by ascending with gradients
       \vspace{-0.5em}$$\Delta_{\omega_i} = \hat{\mathbb{E}}_{\chi_i}[\nabla_{\omega_i} D_{\omega_i}(s, a)] - \hat{\mathbb{E}}_{\chi_E}[\nabla_{\omega_i}D_{\omega_i}(s, a)]$$
       \STATE Clip the weights of $\omega_{i+1}$ to $[-0.01, 0.01]$.
       \STATE Update $\psi_i$ to $\psi_{i+1}$ by descending with gradients
       \vspace{-0.5em}$$\Delta_{\psi_i} = -\lambda_1 \hat{\mathbb{E}}_{\chi_i} [\nabla_{\psi_i} \log Q_{\psi_i}(c|s, a)]$$
       \STATE Take a policy step from $\theta_i$ to $\theta_{i+1}$, using the TRPO update rule with the following objective (without reward augmentation): 
       \vspace{-0.5em}
       $$\hat{\bb{E}}_{\chi_i} [D_{\omega_{i+1}}(s, a)] - \lambda_1 L_I(\pi_{\theta_i}, Q_{\psi_{i+1}}) - \lambda_2 H(\pi_{\theta_i})$$ 
       \STATE or (with reward augmentation):
       \vspace{-0.5em}$$\hat{\bb{E}}_{\chi_i} [D_{\omega_{i+1}}(s, a)] - \lambda_0 \eta(\pi_{\theta_i}) - \lambda_1 L_I(\pi_{\theta_i}, Q_{\psi_{i+1}}) - \lambda_2 H(\pi_{\theta_i})$$ 
%Specifically, take a KL-constrained natural gradient step with
%        \STATE
%            \begin{equation}
%                \hat{\mathbb{E}}_{\tau_i}[\nabla_\theta\log{(\pi_\theta(a|s))}Q(s, a)],
%            \end{equation}
%        \STATE where $Q(\bar{s}, \bar{a})=\hat{\mathbb{E}_{\tau_i}}[D_{\tau_{i+1}}(s, a)|s_0=\bar{s}, a_0=\bar{a}]$ \js{reuse of notation}
   \ENDFOR
\end{algorithmic}
\end{algorithm}

\section{Visual Illustrations of the Learned Policies}
\label{sec:videos}
A video showing the experimental results is available at \href{https://youtu.be/YtNPBAW6h5k}{\texttt{https://youtu.be/YtNPBAW6h5k}}.

\subsection{InfoGAIL Policies}

\begin{figure*}[h!]
  \centering
  \includegraphics[width=\textwidth]{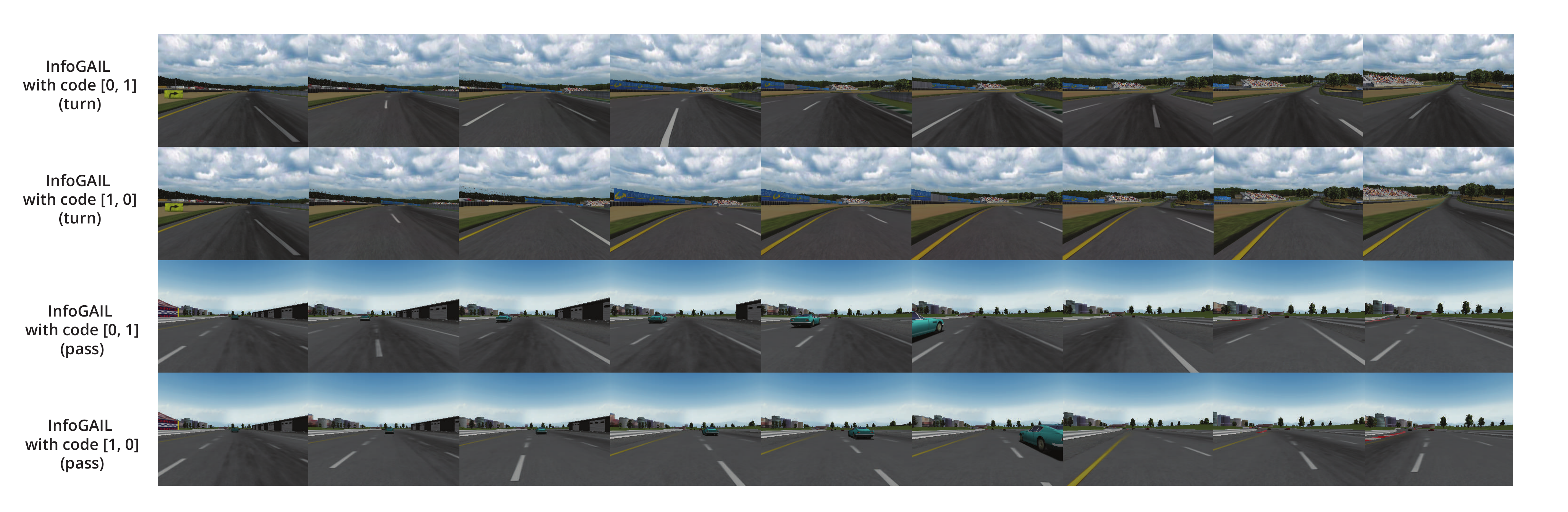}
  \caption{
  {\bf Visual inputs for the InfoGAIL policies in \textit{pass} and \textit{turn} experiments.} The latent code controls the behavior of the policies. %Videos are included in the supplementary material.
  }
  \label{fig:infogail}
 \end{figure*}

Figure~\ref{fig:infogail} provides an illustration of the learned policies under different latent codes.

%Full video is available in our supplemented video.

% For turn experiments:
% \begin{itemize}
%  \item \textbf{turn\_inner.mp4} denotes the agent takes a turn at the inner lane;
%  \item \textbf{turn\_outer.mp4} denotes the agent takes a turn at the outer lane;
%  \item \textbf{turn\_middle.mp4} denotes the policy under the latent code $[0.5, 0.5]$. Interestingly, the agent takes a turn at the middle lane, which suggests that it might be able to generalize (to some extent) to cases where no trajectory is provided.
%  \end{itemize}

% For pass experiments:
% \begin{itemize}
%  \item \textbf{pass\_left.mp4} denotes the agent passes the other car from the left;
%  \item \textbf{pass\_right.mp4} denotes the agent passes the other car from the right.
% \end{itemize}

\subsection{InfoGAIL Policies for Ablation Experiments}

\begin{figure*}[h]
  \centering
  \includegraphics[width=\textwidth]{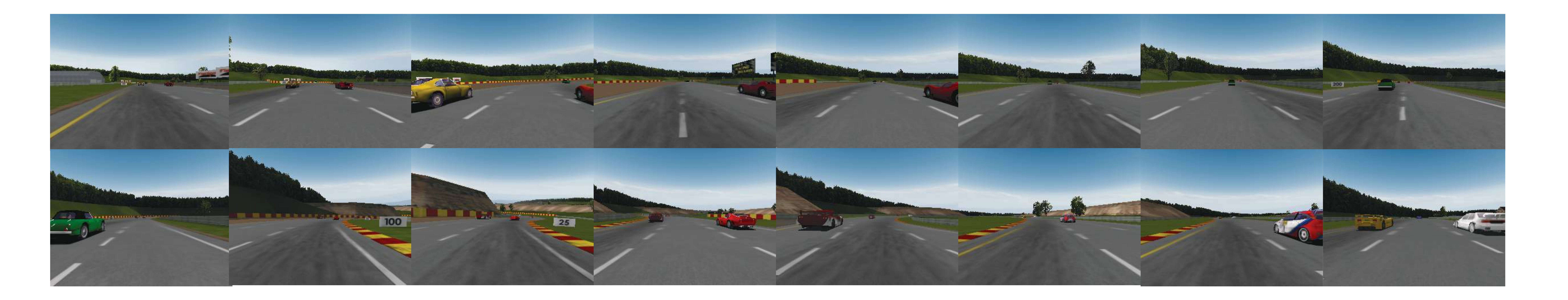}
  \caption{\textbf{Visual inputs for our policy in the ablation experiment.} During the rollout, our policy is able to pass several other cars. A video is included in the supplemented video.}
  \label{fig:gail_pixel}
 \end{figure*}

In all the experiments, the agent is designed to drive at a speed around 80 km/h; therefore the control has to be very precise. 

\subsection{Failure Case for Reward Augmentation}
We also provide an additional video, which demonstrates a failure case of reward augmentation: the failure is caused by assigning a excessively high reward for avoiding collision. Here the agent is too "afraid" of the previous car, and tries to use repeated turning to increase friction and slow down dramatically, which is clearly undesirable behavior.

\end{document}